\documentclass[a4paper,12pt]{article}

\usepackage{times}
\usepackage[german,english]{babel}
\usepackage[latin1]{inputenc}

\usepackage[a4paper]{anysize}
\marginsize{3cm}{3cm}{1.0cm}{2.0cm}

\usepackage[round,colon]{natbib}

\usepackage{graphicx}
\usepackage{tabularx}

\usepackage{amsmath}
\usepackage{amsfonts}
\usepackage{amssymb}

\usepackage{caption}
\usepackage[font=footnotesize]{subfig}

\newcommand{\vecs}{{\bf s}}
\newcommand{\vecx}{{\bf x}}
\newcommand{\vecbeta}{\boldsymbol{\beta}}

\newcommand{\rb}[2]{\raisebox{-#1}[#1]{#2}}

\newcommand{\mf}{\small}

\newcommand{\iw}{w_{\rm IM}}
\newcommand{\ih}{h_{\rm IM}}

\newcolumntype{R}{>{\centering\arraybackslash}X}

\newdimen\arrayruleHwidth
\makeatletter
\def\Hline{\noalign{\ifnum0=`}\fi\hrule \@height \arrayruleHwidth
\futurelet \@tempa\@xhline}
\makeatother

\begin{document}

\title{Detecting Affordances by Visuomotor Simulation}

\author{Wolfram Schenck$^{1,2}$, Hendrik Hasenbein$^{2}$, Ralf M\"oller$^{2,3}$ \\
\\
  $^1$ Faculty of Engineering and Mathematics, \\[-0.13cm]
  Bielefeld University of Applied Sciences, Bielefeld, Germany \\
  $^2$ Computer Engineering Group, Faculty of Technology, \\[-0.13cm]
  Bielefeld University, Bielefeld, Germany \\
  $^3$ CITEC --- Center of Excellence Cognitive Interaction Technology, \\[-0.13cm]
  Bielefeld University, Bielefeld, Germany \\
{\tt wolfram.schenck@fh-bielefeld.de}}

\date{}

\maketitle

\begin{abstract}
The term ``affordance'' denotes the behavioral meaning of objects. We propose a cognitive architecture for the detection of affordances in the visual modality. This model is based on the internal simulation of movement sequences. For each movement step, the resulting sensory state is predicted by a forward model, which in turn triggers the generation of a new (simulated) motor command by an inverse model. Thus, a series of mental images in the sensory and in the motor domain is evoked. Starting from a real sensory state, a large number of such sequences is simulated in parallel. Final affordance detection is based on the generated motor commands. We apply this model to a real--world mobile robot which is faced with obstacle arrangements some of which are passable (corridor) and some of which are not (dead ends). The robot's task is to detect the right affordance (``pass--through--able'' or ``non--pass--through--able''). The required internal models are acquired in a hierarchical training process. Afterwards, the robotic agent is able to distinguish reliably between corridors and dead ends. This real--world result enhances the validity of the proposed mental simulation approach. In addition, we compare several key factors in the simulation process regarding performance and efficiency.
\end{abstract}

\vspace{3cm}

{\small \noindent {\bf Funding statement:} This research received no specific grant from any funding agency in the public, commercial, or not-for-profit sectors.}

\clearpage

\section{Introduction}

The term ``mental imagery'' describes the process of activating sensory and/or motor representations within the cognitive system of an agent without actual sensory inflow and without actually executing motor actions. Introspective experience suggests that humans are able to generate mental images, and neuroimaging studies have shown that these mental images have neural correlates in the corresponding cortical areas which are usually involved in visual and motor processing \cite[]{jeannerod+01,kosslyn+93}.
One commonly accepted function of mental imagery is the mental practice of movement tasks, e.g. in sports \cite[]{martin+99} or rehabilitation \cite[]{jackson+01}.

We suggest that mental imagery --- at the subconscious level --- could be the basis for the detection of affordances \cite[]{gibson79} in visual perception. Affordances are the behavior--related properties of entities in an agent's environment; for example, surfaces can be ``stand-on-able'', ``climb-on-able'', or ``sit-on-able''. Accordings to Gibson \cite[]{gibson79,norman02}, visual perception means to directly perceive these behavioral meanings by picking up certain ``invariants'' from the visual information, and through active exploration of the environment without preceding object recognition. Quite the contrary, the object is ``recognized'' by its affordances. Our approach closely follows these lines of thought, however, the active exploration of the environment is replaced by an internal simulation of movement sequences in which mental images of sensory states and motor commands are generated:
The sensory outcome in these simulated sequences is evaluated to guide the simulation process, and a final set of movement sequences serves to determine the behavioral meaning.
In the end, object recognition is based \emph{on the internally generated motor commands} (and optionally on the evaluation of the predicted sensory states) and not on sensory features (even though low--level features may be used in the simulation).
This approach has been termed ``perception through anticipation'' (PtA) in previous publications \cite[e.g.,][]{moeller99}. It replaces the rather vague statements of Gibson's theory on how affordances are detected by a clearly described computional mechanism.

During the last decade several ``simulation theories'' were proposed in cognitive science (and related areas) which basically state that simulation is an important foundation for perception and higher--level cognition \cite[]{hesslow02,cruse03b,wolpert+03,holland+03,grush04,pezzulo+09}. For example, \cite{hesslow02} argued that thinking consists of a simulated interaction with the environment, and \cite{grush04} suggested the ``emulation theory of representation'' which links imagery and perception to motor simulation \cite[see also][]{jeannerod+01}. \cite{marques+09} analyzed the necessary and sufficient requirements for cognitive architectures which are based on internal simulation: Such systems have to be capable of holding covert motor states and internally generated sensory states (i.e., mental imagery), they have to be able to predict future sensory states, and they require goals and mechanisms for the evaluation of sensory states and for action selection \cite[see also][]{hesslow02}. In consequence, the whole simulation approach rests on the assumption that the human cognitive system fulfills these preconditions, and any computational model from this area has to meet these requirements. This is the case for the PtA approach. Furthermore, current neurophysiological and psychological studies and computational models provide converging evidence that the brain is actually an ubiquitous predictor of future states \cite[]{bar07,schubotz07,butz+08d}.

In the present study, a robotic agent, a mobile robot with omnidirectional camera (see Fig.~\ref{fig_pio}), learns to distinguish between two types of obstacle arrangements: dead ends and corridors. One example for a dead end and one for a corridor are shown in  Fig.~\ref{fig_compare}. It is important to note that the obstacle arrangement as a whole is interpreted here as one single entity whose behavioral meaning has to be uncovered, while each obstacle on its own serves only as a low--level feature. The behavioral meanings would be ``pass--through--able'' for a corridor and ``non--pass--through--able'' for a dead end. Another important feature of the presented cognitive architecture (beyond internal simulation and mental imagery) is the decomposition into sub--models which are aquired through sensorimotor learning in a hierarchical way: visual and tactile ``forward models'' for the prediction of sensory states and an ``inverse model'' for the generation of motor commands.

Linking perception to sensorimotor simulation has two additional interesting implications \cite[]{hoffmann07}: First, the observer's body size and behavior determine what is perceived in which way. For example, distance is not understood as a metrical measurement but as movement effort \cite[]{witt+08,schenck09a}. In the context of dead end recognition, a dead end is defined as an arrangement of obstacles that cannot be passed through \cite[]{moeller99}. This definition implies that it depends on the size of the observer's body what is perceived as dead end and what as corridor. And second, sensorimotor simulation allows for viewpoint invariance: A dead end is recognized by its behavioral meaning and independent from the observer's perspective \cite[]{moeller99}.

Related studies on the detection of affordances have been carried out by various research groups on mobile robots and with robot arm setups \cite[e.g.,][]{fitzpatrick+03, dogar+07, montesano+08, stoytchev08}. Iterative sensory anticipation on (simulated) robots has been studied among others by \cite{tani+99} and \cite{ziemke+05}.
The studies by \cite{gross+99} and \cite{hoffmann07} are especially close to our own work: \cite{gross+99} trained a robotic agent to predict the optical flow caused by self--motion, and by this predictive ability the system could generate collision-free movement sequences. The mobile robot system developed by \cite{hoffmann07} could even predict how the whole image of an obstacle arrangement would look like after a movement (although these images were of rather small size). Based on this ability, dead ends and corridors were distinguished by internal simulation. Our own work goes beyond these approaches by using more complex scenarios, a hierarchical learning approach, and by putting a strong emphasis on the question of how to generate movement sequences during internal simulation in an effective way.

Our cognitive architecture has already been tested in a pure simulation study \cite[]{moeller+08}. The achievement of the present work is to demonstrate that dead end recognition in the real world with a real robot setup can be successfully implemented based on the same principles, enhancing the validity of the overall approach. Furthermore, we extend our previous work by a more thorough experimental analysis in which we consider, for example, different types of inverse models.

The remainder of this article is organized as follows: In Sect.~\ref{sect_setup}, robot setup and image processing are described, Sect.~\ref{sect_comp} explains the computational model and its components, Sect.~\ref{sect_exp} the experimental study and its results, and Sect.~\ref{sect_concl} contains the discussion.

\section{Setup and Image Processing}\label{sect_setup}

\subsection{Setup}

\begin{figure}[tb]
\begin{center}
\includegraphics[width=1.0\textwidth]{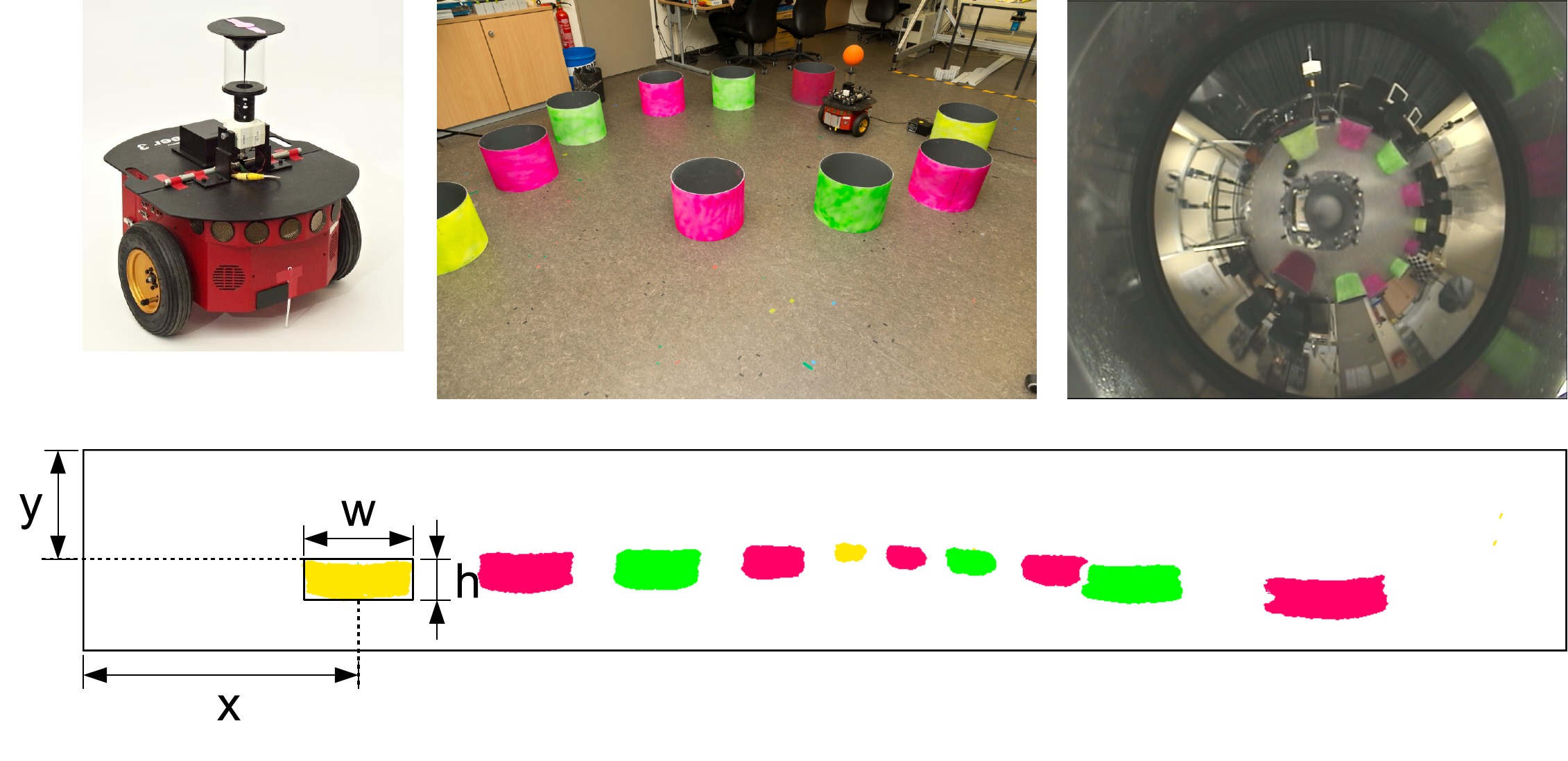}
\caption{\label{fig_pio}Upper left: The robotic agent, a mobile Pioneer 3DX robot equipped with an omnidirectional camera. Upper middle: Obstacle arrangement (external view). Upper right: The same obstacle arrangement as recorded by the robot's camera. Lower image: Detected obstacles after image unfolding and color segmentation. Each obstacle is characterized by its width ($w$), height ($h$), horizontal position ($x$), and its vertical offset from the upper border of the unfolded panoramic view ($y$).}
\end{center}
\end{figure}

\begin{figure}[tb]
\begin{center}
\begin{tabular}{cc|cc}
\large Dead end & \hspace{0.15cm} & \hspace{0.15cm} & \large Corridor \\[0.2cm]
\includegraphics[width=0.4\textwidth]{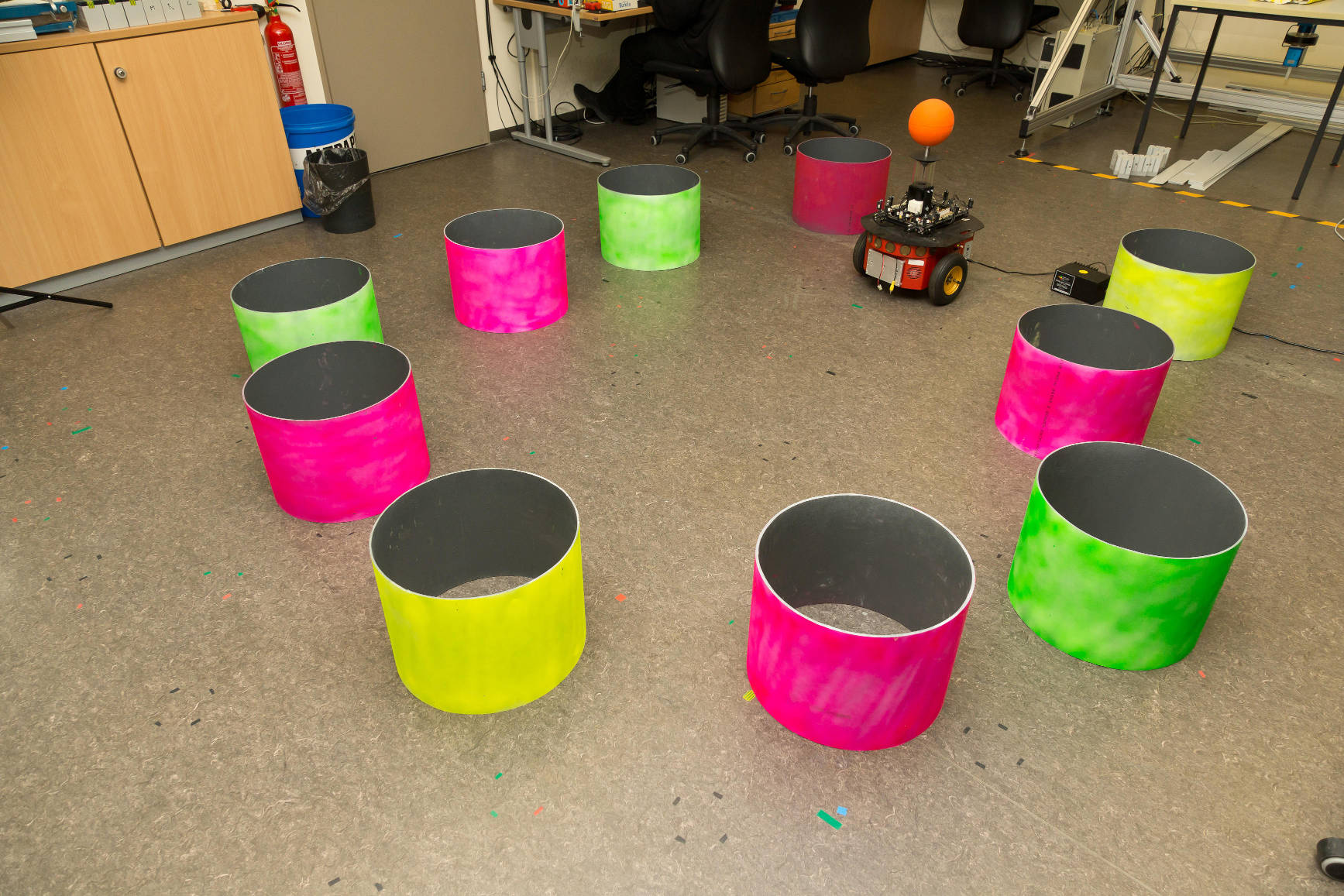} & && \includegraphics[width=0.4\textwidth]{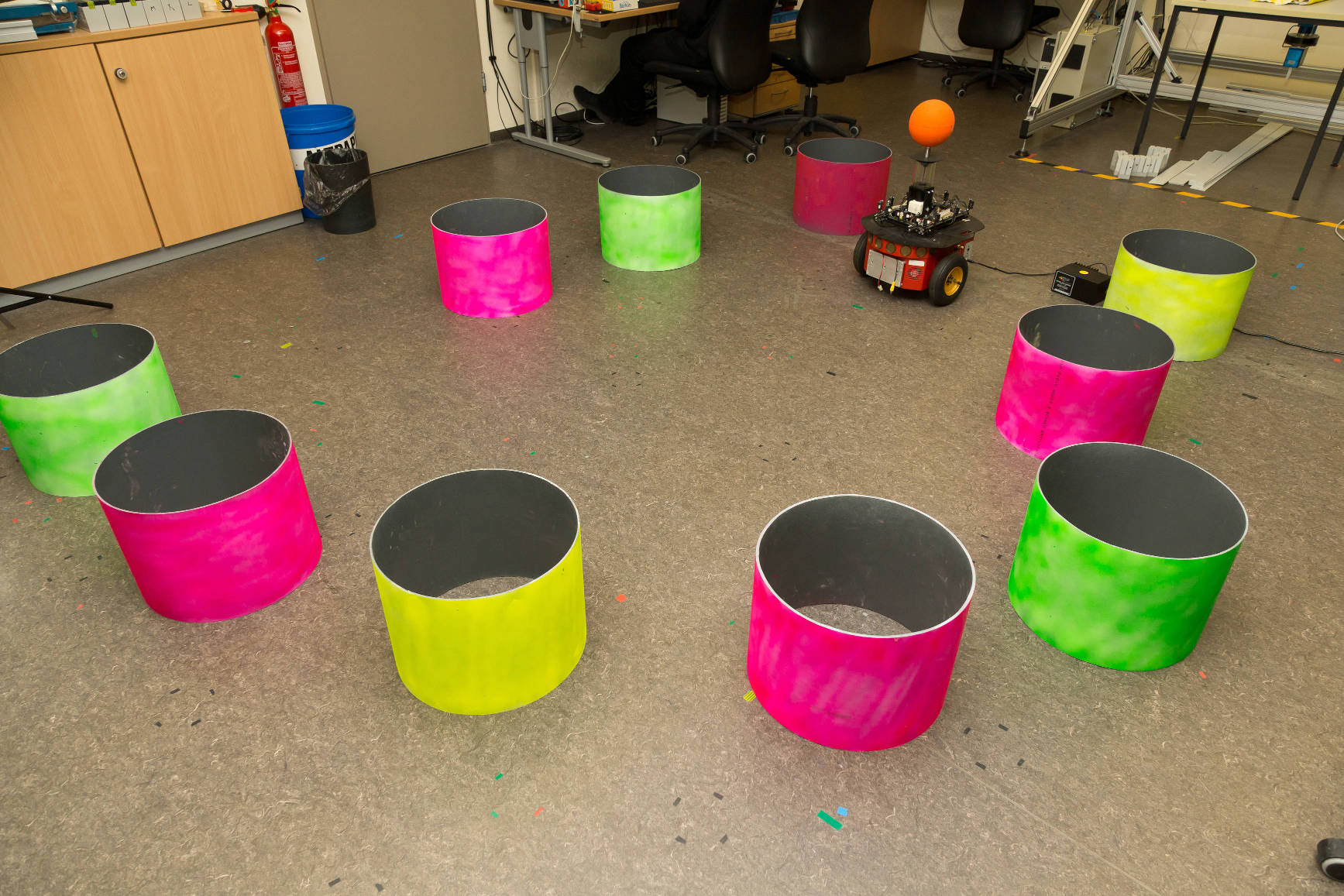} \\[0.2cm]
\includegraphics[width=0.4\textwidth]{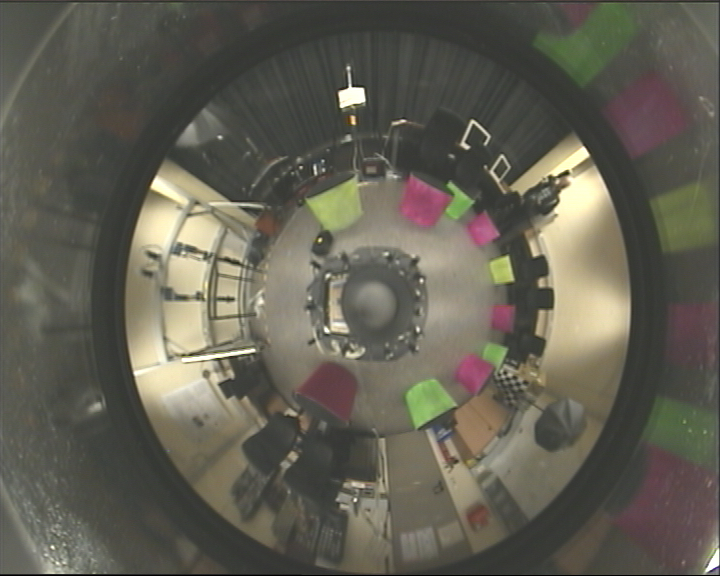} & && \includegraphics[width=0.4\textwidth]{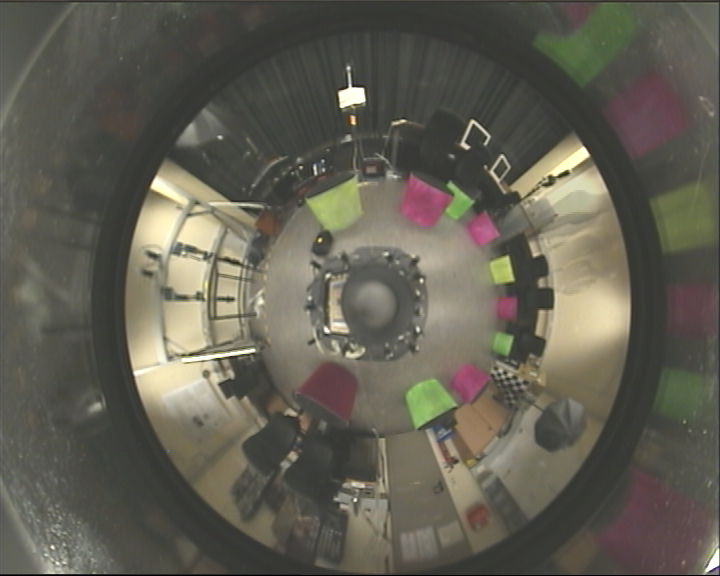} \\[0.2cm]
\includegraphics[width=0.4\textwidth]{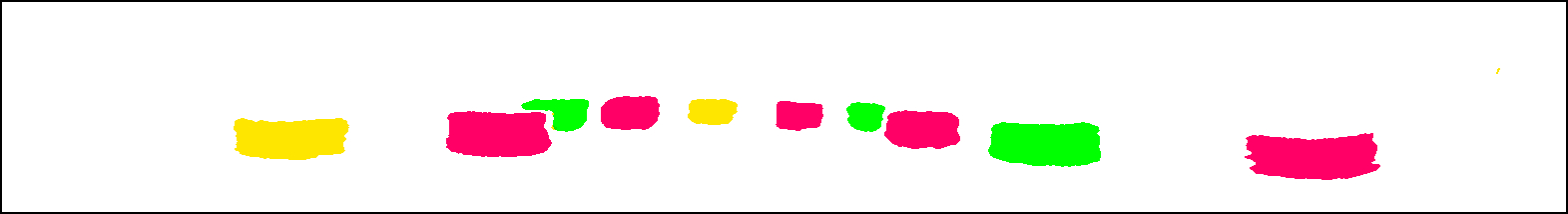} & && \includegraphics[width=0.4\textwidth]{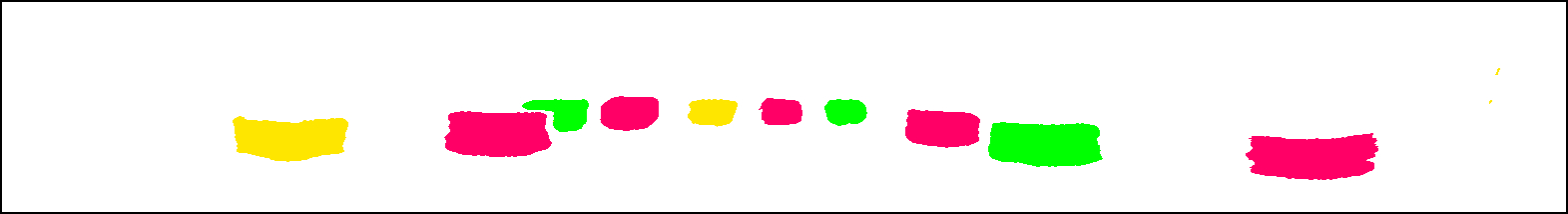} \\[0.2cm]
\end{tabular}
\caption{\label{fig_compare}Comparison between a dead end (left column) and a corridor (right column). In the top row, an external view of the robot and the obstacle arrangement is shown, in the second row the unprocessed images from the robot's omnidirectional camera, and in the third row the corresponding result of image unfolding and color segmentation. The center in the unfolded images marks the direction of forward movements of the robot.
}
\end{center}
\end{figure}

The mobile robot in this study is based on the Pioneer 3DX platform. The mounted camera records images from a mirror surface which reflects a complete 360$^\circ$--view of the environment (Fig.~\ref{fig_pio}). These images are unfolded to show a horizontal panoramic view. Afterwards, the obstacles within an arrangement are detected through color segmentation and some filtering rules (second row in Fig.~\ref{fig_pio}). The obstacle colors are either yellow, green, or red. The obstacles have a diameter of 40~cm (roughly the same as the robot) and a height of 30~cm. The obstacle arrangements can extend over an area up to the size of $3\times4$~m. Beyond this distance, the resolution of the camera is not sufficient for reliable obstacle identification. An exemplary obstacle arrangement in our laboratory is shown in the upper middle of Fig.~\ref{fig_pio}.

The final sensory features are the width, the horizontal position, and the vertical offset of each obstacle $i$: $\vecs_{i,t} = \left(w_{i,t},x_{i,t},y_{i,t}\right)$ in time/simulation step $t$. Furthermore, the height $h_{i}$ of each obstacle $i$ is recorded in the beginning of movement sequences for the correction of the sensory state (see Sect.~\ref{sect_improc_corr}). The overall sensory state $\vecs_{t}$ is a collection of the vectors $\vecs_{i,t}$ for all detected obstacles. For the movements $m_t$ the robot has three different commands available: It can either move forward by ca.~10~cm, or it can rotate left or right by ca.~15 degrees.

\subsection{Image Processing}\label{sect_improc}

\begin{figure}[tb]
\begin{center}
\includegraphics[width=1.0\textwidth]{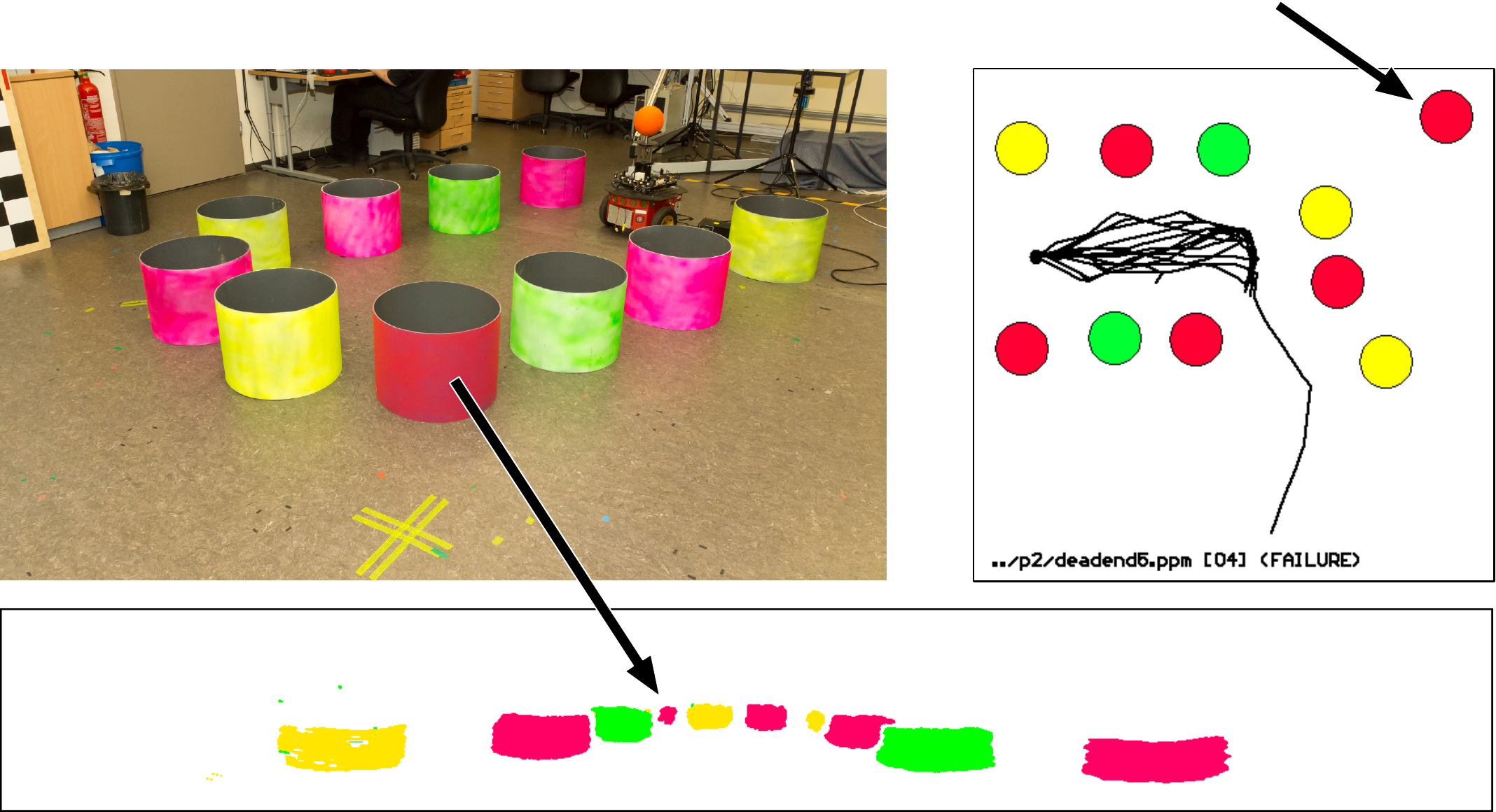}
\caption{\label{fig_deadendMiss}Partly occluded obstacles can be mislocated because their width is not correctly detected. In this dead end, this is the case for the red obstacle marked with an arrow and for the yellow obstacle vis--\`{a}--vis (further explanation in the main text).}
\end{center}
\end{figure}

\subsubsection{General Processing}

The camera images of the robot have PAL resolution (720$\times$576 pixels). They are slightly denoised and unfolded by a polar mapping to a horizontal panoramic view with a size of $\iw \times \ih$ pixels (with $\iw=1571$ and $\ih=214$). In this horizontal panoramic view, the obstacles are identified by color segmentation based on saturation and hue in the HSV color space. During segmentation, the leftmost and rightmost image columns are handled as if they were directly neighboured, i.~e.~the panoramic image is closed horizontally. All segments below the horizon, with a minimum fill ratio (number of colored pixels devided by area of bounding box larger than 0.4), and with a minimum area (150 pixels) are included in the set of detected obstacles.

\subsubsection{Correction of Sensory States}\label{sect_improc_corr}

Specific problems arise if obstacles partly occlude each other. In theory, the upper part of the rear obstacle should be fully visible because the height of the obstacles was chosen such that they appear completely below the horizon in the panoramic images. However, for far--away objects this is not always the case: If two obstacles are close to each other, the upper part of the rear obstacle might cover only one row of pixels which is easily destroyed during denoising and color segmentation. In such a case, the rear obstacle appears as if it was much slimmer than it is in reality. This is illustrated in Fig.~\ref{fig_deadendMiss}. The red obstacle marked with an arrow appears far too small in the processed image. Because the obstacle width is an important indicator of its distance to the robot, this is a fatal misperception. For this reason, we incorporated a correction step which adjusts $w_{i}$ based on $y_{i}$ and $h_{i}$ because these three parameters encode the distance between robot and obstacle in a redundant way.

\begin{figure}[tb]
\begin{center}
\begin{tabular}{cc}
\multicolumn{2}{c}{\includegraphics[width=0.45\textwidth]{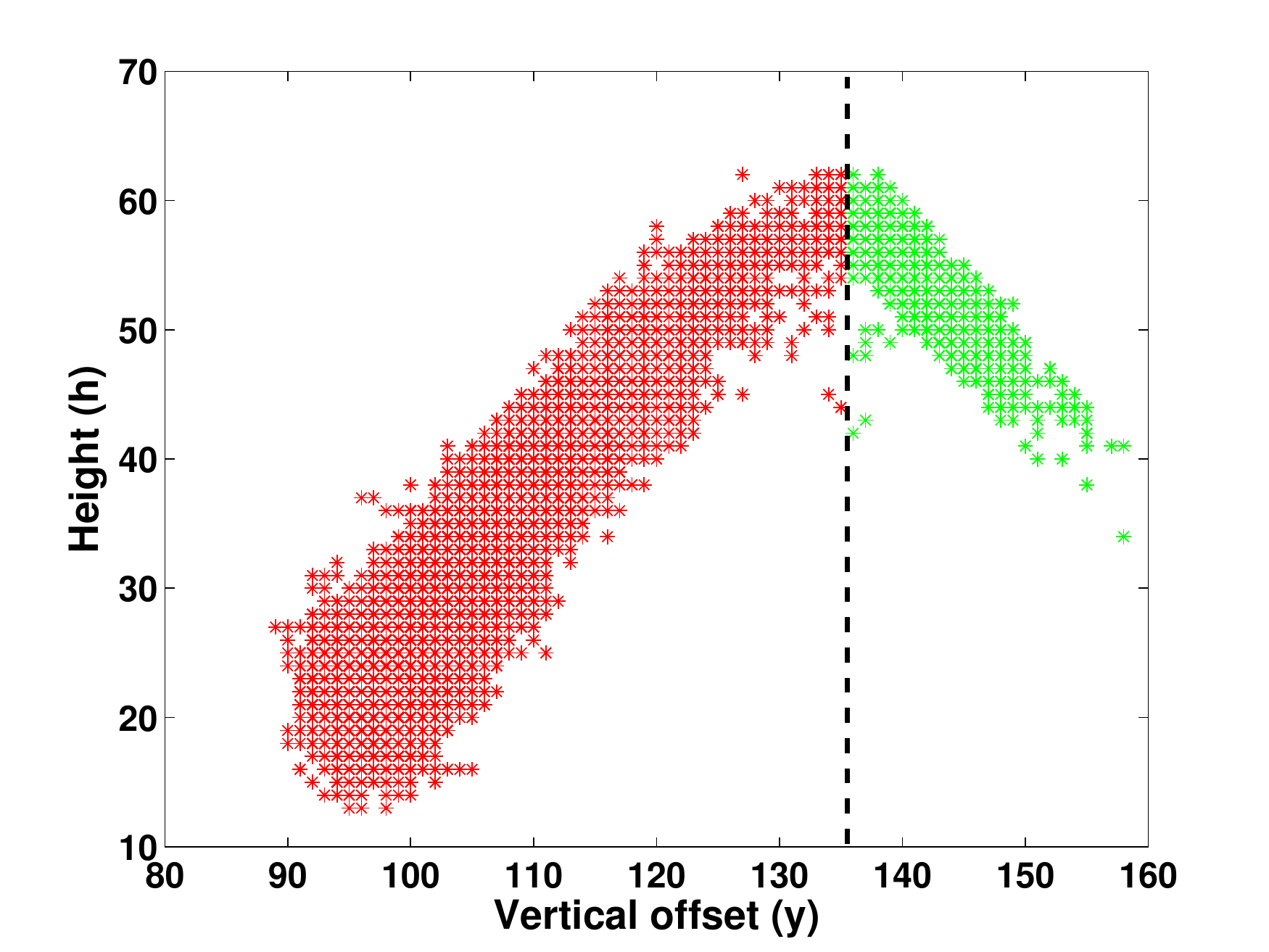}} \\
\includegraphics[width=0.45\textwidth]{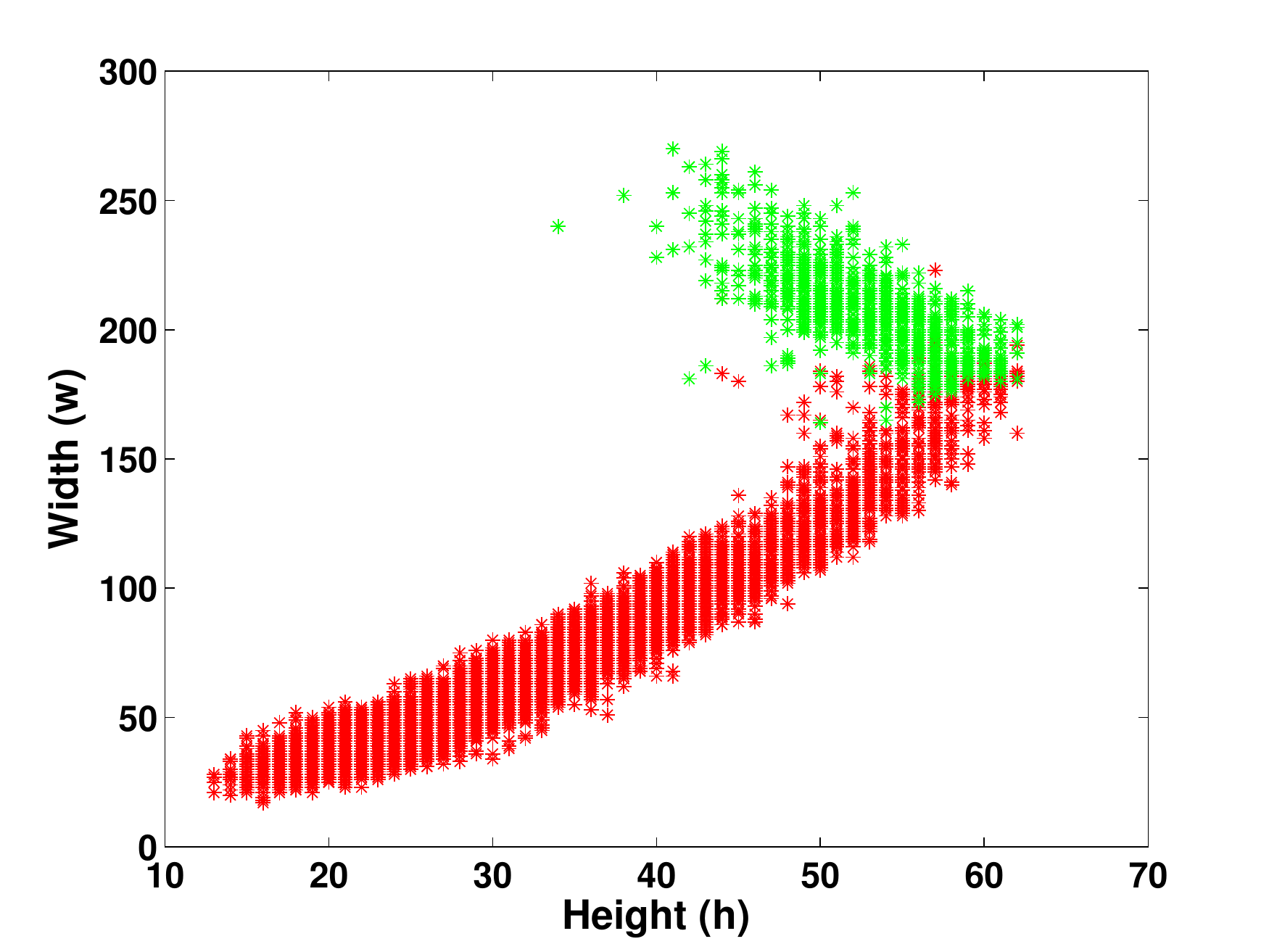} &
\includegraphics[width=0.45\textwidth]{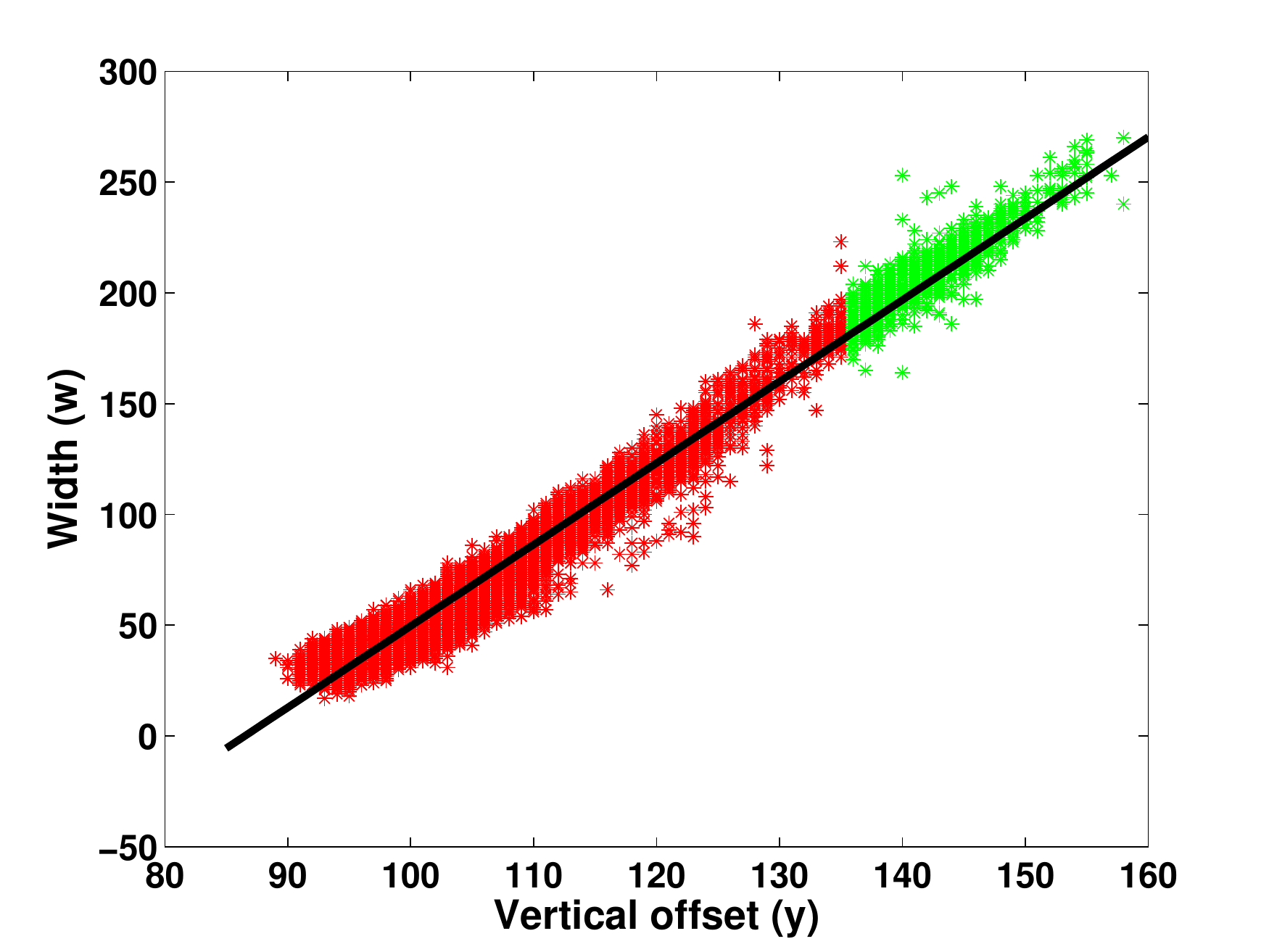} \\
\end{tabular}
\caption{\label{fig_obstacleStat} Relationships between the obstacle characteristics vertical offset ($y$), width ($w$), and height ($h$). $13,000$ data points are shown in each figure. The vertical dotted line in the topmost figure marks the border between obstacles with $y \le 135$ (red) and $y > 135$ (green). This color assignment is kept in the two figures below. The figure on the lower right shows in addition a linear function which is fitted to the data: $w = -316.2 + 3.68 y$ (this is the inverse to Eqn.~\ref{eqn_fity}).}
\end{center}
\end{figure}

This correction is based on an analysis of the images which were collected as training data for the visual forward model (see Sect.~\ref{sect_fm}). Overall, $13,000$ obstacles were recorded in these images, all of them completely visible; color and position relative to the robot were partly systematically, partly randomly varied. Fig.~\ref{fig_obstacleStat} depicts the relationships between vertical offset $y$, height $h$, and width $w$ for the corresponding $13,000$ segments. First of all, the data is devided into two parts based on the relationship between $y$ und $h$ (uppermost plot in Fig.~\ref{fig_obstacleStat}): Segments with small $y$ values ($y \le 135$, red dots) are far away from the robot's chassis and fully visible while segments with large $y$ values ($y > 135$, green dots) are close to the robot and partly hidden by its chassis. For this reason, with increasing $y$ values (robot gets closer) $h$ increases first and then starts to decrease as soon as the robot's chassis begins to hide part of the obstacle.

For each part of the data both the relationships between $h$ and $w$ and between $y$ and $w$ (lower left and right plot in Fig.~\ref{fig_obstacleStat}) are close to linear. For this reason, linear models incorporating $y$ and $h$ were fitted to these data points:
\begin{eqnarray*}
\widehat{w} &=& 2.12 y + 1.35 h - 197.6 \; , \; y \le 135 \quad (R^2=0.97)\\
\widehat{w} &=& 3.53 y - 0.039 h - 291.4 \; , \; y > 135 \quad (R^2=0.80)
\end{eqnarray*}
Based on these models, the corrected width value $\widetilde{w}$ is computed by $\widetilde{w} = (w+\widehat{w})/2$, giving half weight to the original value and half weight to the estimation based on $y$ and $h$.
Afterwards, the $y$ value is adjusted to match the new width value $\widetilde{w}$:
\begin{equation}
\widetilde{y} = 0.265 \widetilde{w} + 87.0 \label{eqn_fity}
\end{equation}
This linear model was obtained from the combined data (see lower right plot in Fig.~\ref{fig_obstacleStat}; $R^2=0.98$). In this way, the corrected sensory state $\widetilde{\vecs}_{i,0} = \left(\widetilde{w}_{i,0},x_{i,0},\widetilde{y}_{i,0}\right)$ for each obstacle $i$ in the first time step of a simulated movement sequence is obtained (directly after acquisition of the initial real camera image). Note that this correction is only applied at the start of long--term simulations (Sect.~\ref{sect_simu}), but neither on the training data for the visual forward model (Sect.~\ref{sect_fm}) nor while generating training data through short--term simulation for the inverse model (Sect.~\ref{sect_im}).

\section{Computational Model}\label{sect_comp}

\subsection{Overview}\label{sect_overview}

The main components of the computational architecture are adaptive forward and inverse models (FMs and IMs). The visual FM predicts for each obstacle the change $\Delta\widehat{\vecs}_{i}$ of its sensory state from the previous sensory state $\vecs_{i,t}$ and a motor command $m_t$. The estimated new sensory state $\widehat{\vecs}_{i,t+1}$ is computed by $\widehat{\vecs}_{i,t+1} = \vecs_{i,t} + \Delta\widehat{\vecs}_{i}$. The overall prediction $\widehat{\vecs}_{t+1}$ is composed from the single predictions $\widehat{\vecs}_{i,t+1}$. The visual FM is acquired by collecting a large amount of training data from random movements of the robot with obstacles placed systematically at different distances.
The tactile FM predicts if a collision with an obstacle would occur given $\vecs_{i,t}$ and the scheduled motor command $m_t$.
The evaluation system of the cognitive architecture takes only the tactile events into account: Collisions are bad, sensory states without collision are good.

The IM generates motor commands $m_t$ in response to a wholistic sensory input. This input is a variation of the image shown in the second row of Fig.~\ref{fig_pio}: All obstacles appear -- regardless of their original color -- as white filled circles on a black background (topmost image in Fig.~\ref{fig_imPLS}). The purpose of this processing step is to enable the generation of these wholistic images during the internal simulation when only the predictions of the visual FM are available and not a real segmented panoramic image. The IM is adapted to a movement strategy which favors forward movements as long as possible and turns in the appropriate direction when obstacles appear in the vicinity of the robot. Training data for the IM is generated by internal short--term simulations based on the already trained FMs, starting from several different real--world situations. These  simulations reveal which movement sequence would result in the least curved trajectory without collisions. The first movement of this sequence is the motor output $m_t$ which is linked to the wholistic sensory state of the initial real--world situation. A large number of such learning examples is used to train the IM. An important conceptual aspect of this procedure is that the predictive abilities of the agent have to be already available when motor training starts, because this training is based on internal simulation and mental imagery.

After both FMs and the IM are trained, long--term internal simulations can be carried out which are guided by the motor commands from the IM. In this iterative long--term simulation, predictions by the FMs generate new sensory states which trigger new motor commands from the IM which in turn lead to new predictions (see Fig.~\ref{fig_sim}). To classify an obstacle arrangement, the robot first processes the real visual input to detect the image features (obstacles) $\vecs_{i,0}$. Starting from $\vecs_{0}$, several long-term simulations are carried out. If the motor commands in any of the generated movement sequences indicate that a passage is possible, this arrangement is classified as corridor, otherwise as dead end. Thus, the final recognition of the behavioral meaning is purely based on the motor commands in the simulated movement sequences.

\subsection{Forward Models}\label{sect_fm}

\subsubsection{Visual Forward Model}

For the visual FM, ca.~$13,000$ learning examples were collected in 120 random motor sequences with an average length of 55 steps. The distance between robot and obstacles and the initial orientation of the robot were varied systematically between sequences. A maximum number of two obstacles was used in each sequence. Each learning example has the structure $\left[\left(\vecs_{i,t},m_t\right) \longrightarrow \Delta\vecs_i \right]$ with $\Delta\vecs_i = \vecs_{i,t+1} - \vecs_{i,t}$. The training set for forward movements consists of ca.~$6300$ learning examples, for left rotations of ca.~$3100$ examples, and for right rotations of ca.~$3600$ examples.

We tested different implementations for the visual FM \cite[e.g., multi--layer perceptron;][]{rumelhart+86}, but in the end the best results were obtained by fitting analytical functions to the training data. For rotations, the output of the visual FM is constant:
\begin{eqnarray*}
\Delta \vecs_{\rm left} &=& \left(\Delta w_{\rm left},\Delta x_{\rm left},\Delta y_{\rm left}\right) = \left(0,60,0\right) \\
\Delta \vecs_{\rm right} &=& \left(\Delta w_{\rm right},\Delta x_{\rm right},\Delta y_{\rm right}\right) = \left(0,-60,0\right)
\end{eqnarray*}
This simple approach already provides a good fit to the data; it means that rotation commands to the robot are executed as nearly pure rotations on the spot. However, the data also shows that the robot does not turn by $\pm15$ degrees as commanded, but just by $\pm13.7$ degrees ($60/\iw \cdot 360^{\circ}\approx 13.7^{\circ}$).

\begin{figure}[tb]
\begin{center}
\begin{tabular}{cc}
\includegraphics[width=0.45\textwidth]{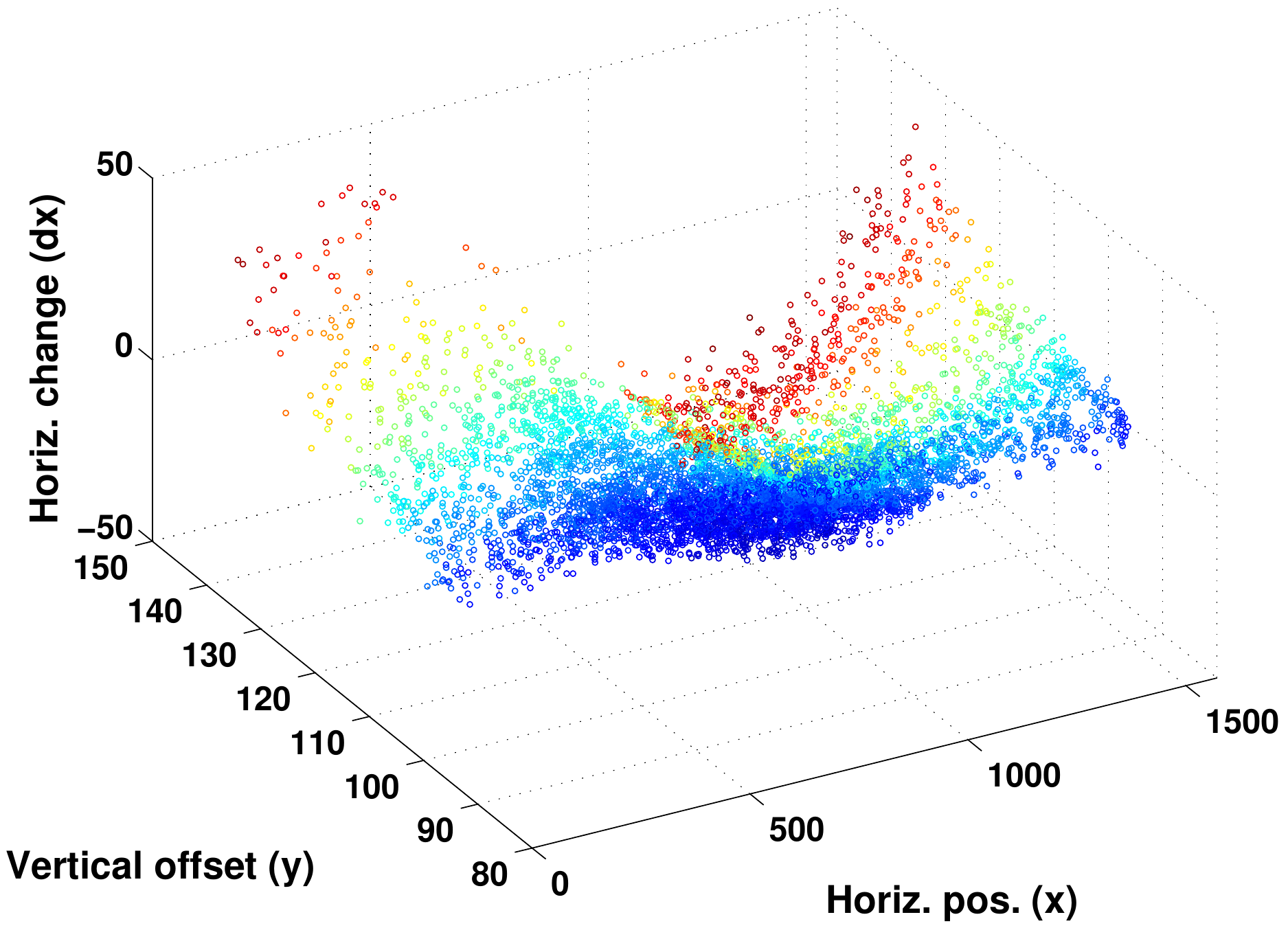} &
\includegraphics[width=0.45\textwidth]{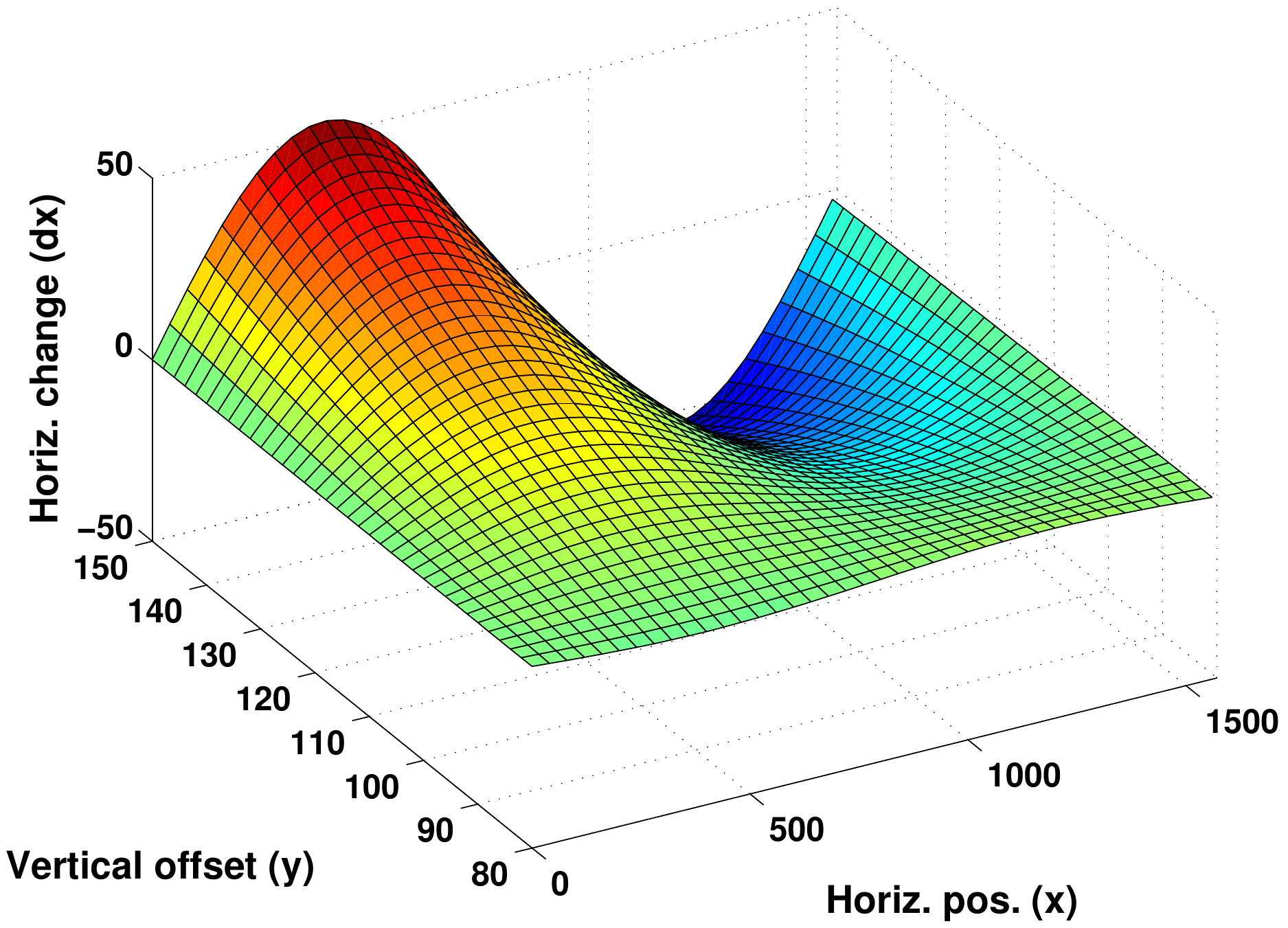} \\[0.5cm]
\includegraphics[width=0.45\textwidth]{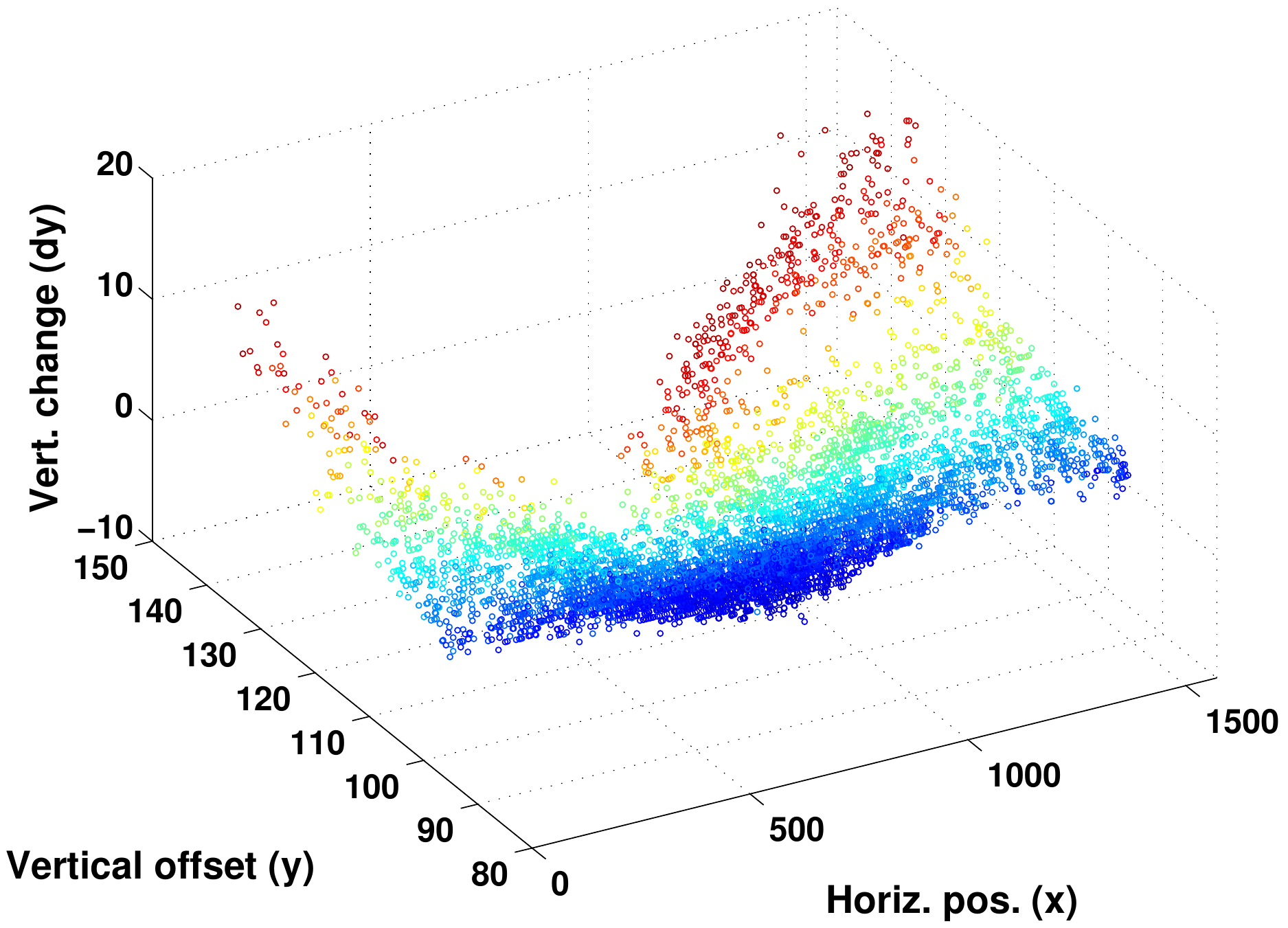} &
\includegraphics[width=0.45\textwidth]{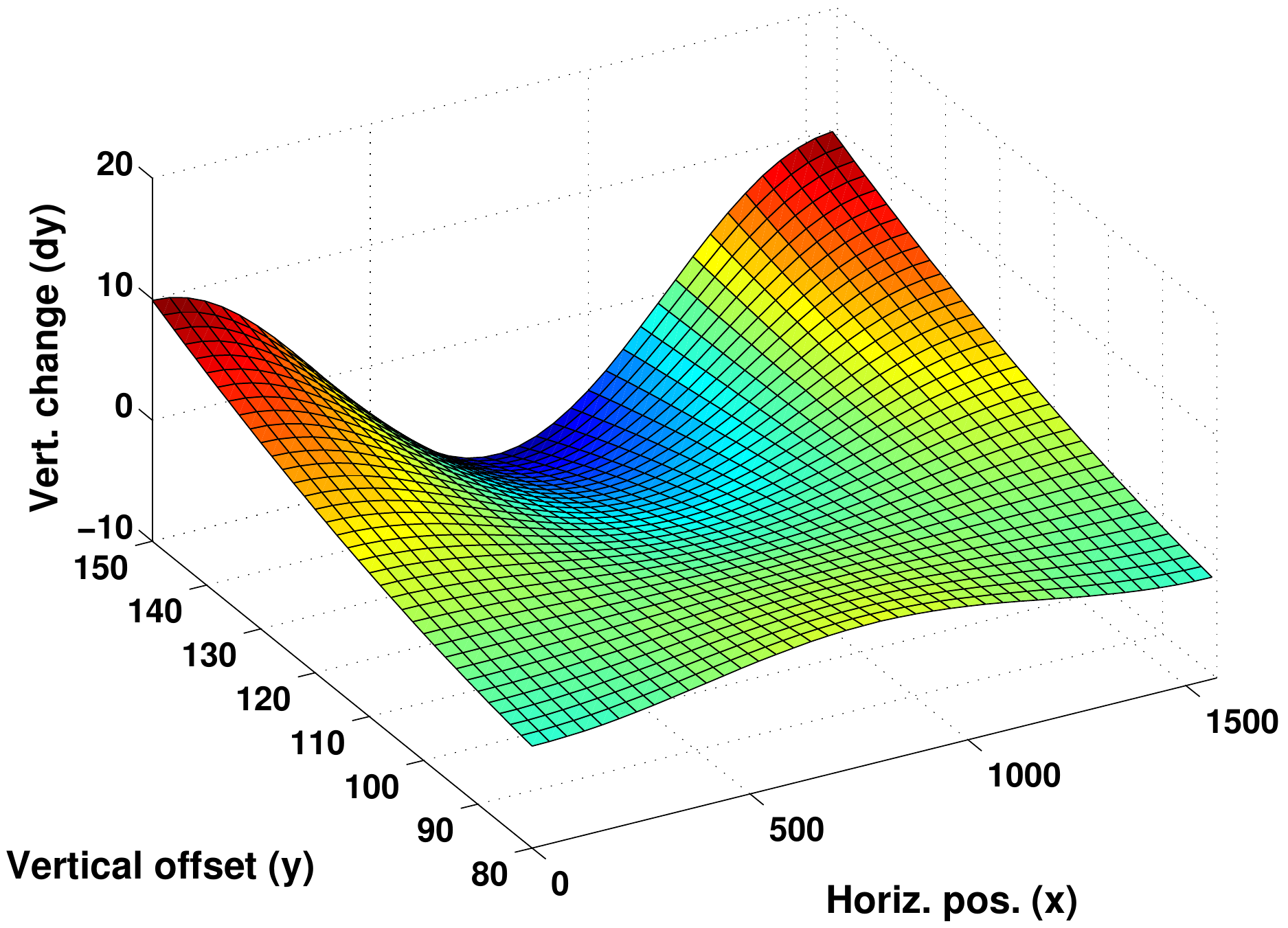} \\
\end{tabular}
\caption{\label{fig_fwmFit}Training data for the visual FM for forward movements on the left, the corresponding fitted functions on the right (see Eqns.~\ref{eqn_dx},\ref{eqn_dy}). The graphs in the top row show $\Delta x_{\rm forw}$ in z--direction, in the bottom row $\Delta y_{\rm forw}$. The coloring only serves to reveal the structure of the 3D data and functions.}
\end{center}
\end{figure}

For forward movements, the following functions were fitted to the data:
\begin{eqnarray}
\Delta w_{\rm forw} &=& cos\left(\frac{2\pi x}{\iw}\right) \left(0.00501 y^2 - 0.482 y\right) \nonumber \\
\Delta x_{\rm forw} &=& sin\left(\frac{2\pi x}{\iw}\right) \left(0.00555 y^2 - 0.473 y\right) \label{eqn_dx} \\
\Delta y_{\rm forw} &=& cos\left(\frac{2\pi x}{\iw}\right) \left(0.00123 y^2 - 0.118 y\right) \label{eqn_dy}
\end{eqnarray}
with $w,x,y$ being components of $\vecs_{i,t}$. Figure \ref{fig_fwmFit} shows the underlying data points and the fitted functions for $\Delta x_{\rm forw}$ and $\Delta y_{\rm forw}$. The output of the visual FM for forward movements is $\Delta \vecs_{\rm forw} = \left(\Delta w_{\rm forw},\Delta x_{\rm forw},\Delta y_{\rm forw}\right)$.
% The new predicted sensory state is computed as $\widehat{\vecs}_{i,t+1} = \left( \dots \right)$.

To test the visual FM, the $120$ random motor sequences for the collection of training data were used again. The FM had to iteratively predict the sensory state for each obstacle in the data (iteratively means here that the input of the FM in each movement step is identical to the output of the FM from the previous step, except for the very first step in which real data is fed into the FM). After 50 movement steps, the result of the iterative prediction was compared to the real sensory state. The average prediction error for $x$ amounted to 35.5 pixels, for $y$ to 4.75 pixels, and for $w$ to 17.0 pixels (for $x$ and $y$, this corresponds to a single--step error of ca.~0.05\% relative to $\iw$/$\ih$, for $w$ to ca.~0.13\% relative to the maximum segment width in the training data). For an iterative prediction over $50$ steps, this is a well acceptable performance.

\subsubsection{Correction Step}

Although the visual FM predicts on average with good precision, it exhibits an undesired extrapolation behavior for far--away obstacles close to the horizon. It was observed in long--term simulations (see Sect.~\ref{sect_simu}) that such objects move away from the robot during forward movements and become larger at the same time. To suppress this unnatural behavior, the predicted width $\widehat{w}$ is used to correct the predicted vertical offset $\widehat{y}$ and vice versa (the obstacle and time step indices are left out here for simplicity). This correction is based on the linear model in (\ref{eqn_fity}) which represents the relationship between $w$ and $y$ for observed obstacles. If the relationship between $\widehat{w}$ and $\widehat{y}$ starts to deviate from this model, this is a clear sign that predictions get into an uncontrolled extrapolation area. To stop this extrapolation, $\widehat{w}$ and $\widehat{y}$ are corrected by the linear model:
\begin{eqnarray*}
\widehat{w}' &=& 3.68 \widehat{y} - 318.2 \quad \mbox{(inverse of (1))} \\
\widehat{w}_{\rm corr} &=& (\widehat{w}+\widehat{w}')/2 \\
\widehat{y}_{\rm corr} &=& 0.265 \widehat{w}_{\rm corr} + 87.0
\end{eqnarray*}
$\widehat{w}_{\rm corr}$ and $\widehat{y}_{\rm corr}$ are the corrected predictions which replace the original ones in further calculations. The original values $\widehat{w}$ and $\widehat{y}$ get the same weight as information sources in the correction process. Note that this additional correction stage is only applied for internal long--term simulations (Sect.~\ref{sect_simu}) and not in any other part of this study.

With enabled correction stage, the average errors after 50 iterative predictions change slightly: The average prediction error for $x$ amounts to 40.6 pixels, for $y$ to 4.16 pixels, and for $w$ to 16.3 pixels. Compared with prediction without correction, the results for $x$ get slightly worse while they improve for $y$ and $w$. The latter improvement seems to be crucial to stabilize long--term simulations.

\subsubsection{Tactile Forward Model}

Because the robot lacks a tactile sensor, collisions are detected by estimating the obstacle distance from the visual data; if an obstacle is too close, this is interpreted as collision. We decided to define a virtual bumper ring around the robot with a diameter of 60 cm. At a distance of 30 cm to the robot's center, obstacles appear in the panoramic image as segments with a width of 215 pixels. For this reason, the tactile FM signals a collision if any prediction $\widehat{w}_{i,t}$ by the visual FM is larger than this value.

\subsection{Inverse Model}\label{sect_im}

\subsubsection{Structure}

As stated in Sect.~\ref{sect_overview}, the IM decides about the motor command $m_t$: forward translation, left turn, or right turn. As input, it receives a variation of the panoramic view: All obstacles appear as white filled circles on a black background (topmost image in Fig.~\ref{fig_imPLS}). Size and position of each circle depends on $\vecs_{i,t}$. The generated image has a size of $197\times42$ pixels, therefore the input data space of the IM has 8274 dimensions.

The IM has two distinct properties: It has to solve a three--fold classification task, and it receives rather high--dimensional input. For this reason, we implemented it by a set of three binary classificators, each of which is a simple linear model (because of the high dimensionality of the data, chances are good that a linear seperation works well). The three binary classification tasks are forward vs.~left, forward vs.~right, and left vs.~right.
The IM contains three regression modules referring to these three classification tasks \cite[see also][pp.~518]{moeller+08}. The first movement $m_1$ in each combination $m_1\mbox{--}m_2$ (with $m_1, m_2 \in \{\mbox{forward}, \mbox{left}, \mbox{right}\}$) is assigned the output value $q=1$ in the regression, the second movement $m_2$ the output value $q=0$. Regression prediction in each module produces an output value $\hat{q}(m_1,m_2)$ that can be used to decide between the two movements. Given an image $\vecx$ (organized as vector), the average of the training images $\overline{\vecx}(m_1,m_2)$, and the regression coefficients $\vecbeta(m_1, m_2)$ for the movement combination $m_1\mbox{--}m_2$, the output value is obtained from %
\begin{equation}\label{eq_regression}
\hat{q}(m_1,m_2)
=
0.5 +  \vecbeta(m_1, m_2)^T \cdot [\vecx - \overline{\vecx}(m_1,m_2)].
\end{equation}
In the reverse direction, we set $\hat{q}(m_2,m_1) = 1 - \hat{q}(m_1,m_2)$. Equation (\ref{eq_regression}) describes a plane for each pair of movements. We can use the plane equation to decide which of the movements $m_1$ and $m_2$ should be executed for a given image $\vecx$.
Now, each of the three alternative actions is assigned a goodness $g(m)$ from
\begin{equation}\label{eq_goodness}
g(m_1) = \min\limits_{{m_2}\atop{m_2 \neq m_1}} \hat{q}(m_1,m_2).
\end{equation}
This operation joins two linear functions by forming a ridge. While equation (\ref{eq_regression}) decides between two movements, the goodness computed in equation (\ref{eq_goodness}) establishes a border between the movement $m_1$ and the two other movements. The movement suggested by the IM is the one with maximal goodness $g$.

\subsubsection{Training}

\begin{figure}[tb]
\begin{center}
\begin{tabular}{lccc}
& \multicolumn{3}{c}{Blob image} \\[0.25cm]
& \multicolumn{3}{c}{\includegraphics[width=0.26\textwidth]{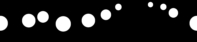}} \\[0.4cm]
& Forward--left & Forward--right & Left--right \\[0.25cm]
\rb{-0.3cm}{Mean} & \includegraphics[width=0.26\textwidth]{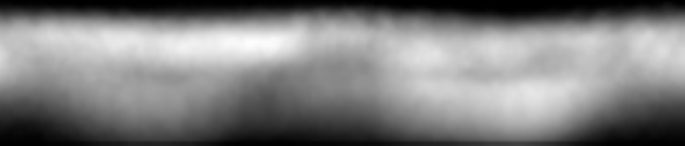} &
\includegraphics[width=0.26\textwidth]{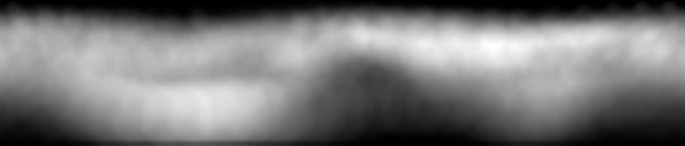} &
\includegraphics[width=0.26\textwidth]{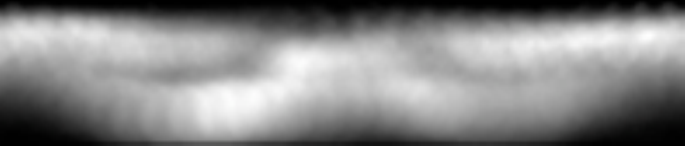} \\[0.15cm]
\rb{-0.3cm}{Beta} & \includegraphics[width=0.26\textwidth]{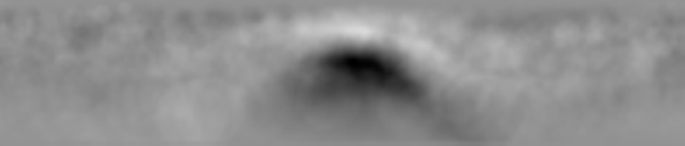} &
\includegraphics[width=0.26\textwidth]{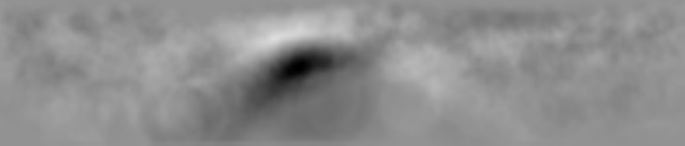} &
\includegraphics[width=0.26\textwidth]{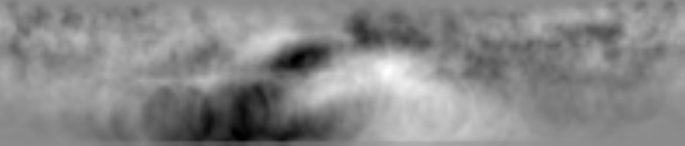}
\end{tabular}
\caption{\label{fig_imPLS}Top: Input image for the inverse model. Bottom: Regression modules of the inverse model. The top row shows the mean of the training data $\overline{\vecx}(m_1, m_2)$ for each module; the correlation coefficients $\vecbeta(m_1, m_2)$ are visualized in the bottom row; see equation (\ref{eq_regression}). Positive values are encoded by white, negative values by black.}
\end{center}
\end{figure}

The conceptual role of the IM in the cognitive architecture is to generate foresighted obstacle avoidance behavior while staying at a straight track as long as possible.
According to this goal, training data for the IM is obtained from an internal simulation process using the already trained visuo-tactile FM \cite[for a detailed description see][pp.~520]{moeller+08}. Short sequences of movements starting from a true image are simulated to find the sequence with minimal costs. The total costs of a sequence contain collision and motor costs. A sequence with collisions can never be the one with minimal costs and is therefore discarded. The motor costs of a simulated sequence are determined from the sum of the costs of each movement in the sequence (forward $0$, rotations $20$), and the sum of the costs when two subsequent movements differ ($10$ when switching from translation to rotation or vice versa, $1000$ when switching between rotations).
The first movement of the best sequence found is stored together with the initial image of this sequence in the training set for the corresponding regression modules. Then the robot performs a real movement and initiates a new search.

Training data for the IM was collected in 6 different arrangements of around 10 obstacles, starting from 8 different viewing directions per arrangement. From each of these 48 different initial situations, a random movement sequence with 100 steps was executed. In each of these overall 4800 movement steps, one learning example for the IM was generated in the way described above (with short--term simulations over 7 steps). For the training of the regression modules, ca.~$1,300$ data points for forward movements, ca.~$2,400$ for left turns, and ca.~$1,100$ for right turns were collected.

Each regression module was adapted to the training data by partial least squares regression (PLS) \cite[]{wold+84}. The resulting mean vectors $\overline{\vecx}$ and regression coefficients $\vecbeta$ are illustrated in Fig.~\ref{fig_imPLS}.
The coefficients $\vecbeta(m_1, m_2)$ found by PLS are plausible and can at least partly be interpreted from the images: For example, the coefficients in the forward-left selector module contain a black region (negative values) close to the center, but slightly misplaced to the right. If an obstacle is present in this region, the module will favor left turns over forward movements.

\subsection{Internal Simulation}\label{sect_simu}

\begin{figure}[tb]
\begin{center}
\includegraphics[width=0.7\textwidth]{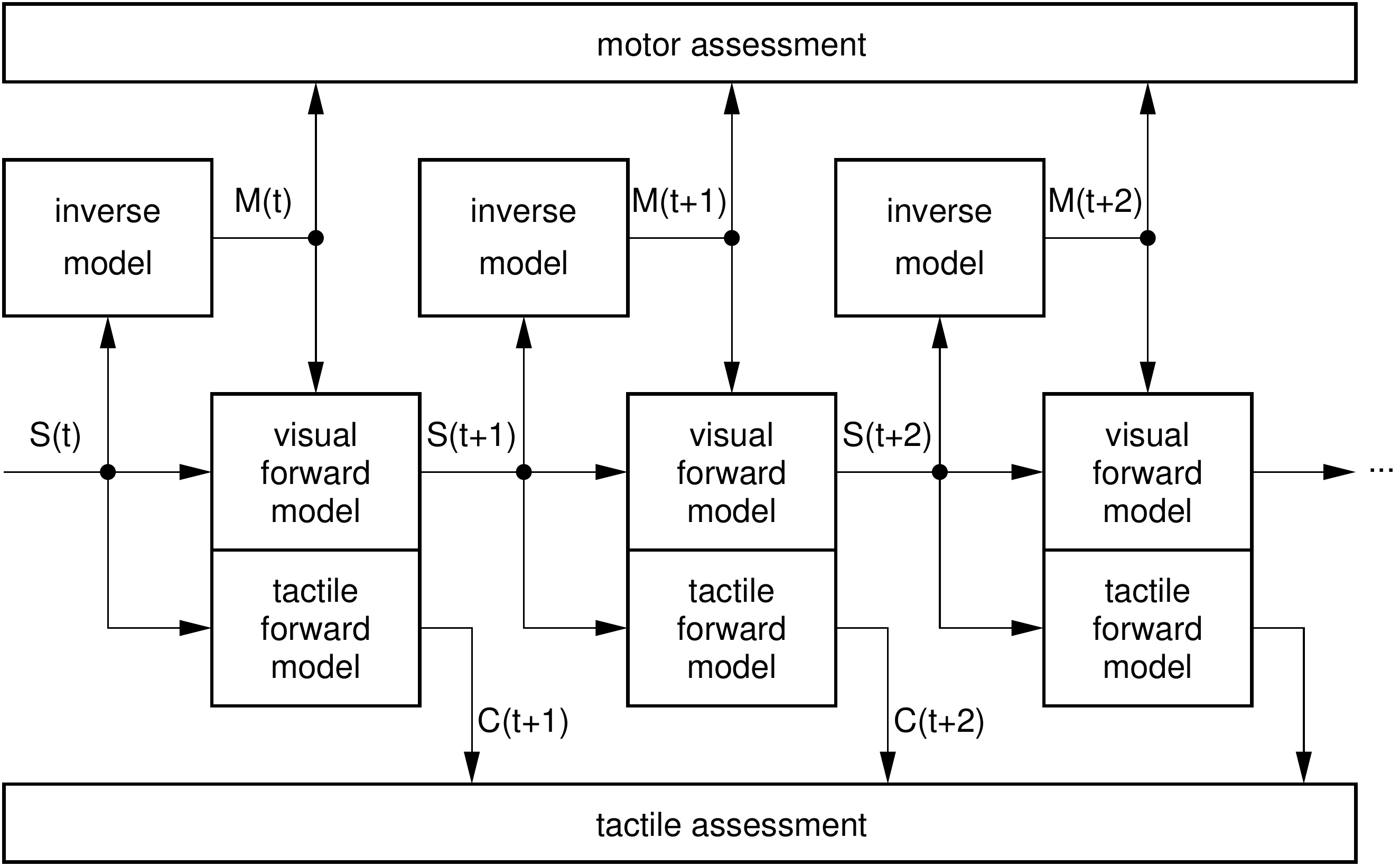}
\caption{\label{fig_sim}Interplay of inverse and forward models in the internal simulation \cite[adapted from][]{moeller+08}. The output of the tactile forward model is denoted as $c_t$ here.}
\end{center}
\end{figure}

The final goal of the cognitive architecture is to reveal by sensorimotor simulation if an obstacle arrangement is ``pass--through--able'' (corridor) or ``non--pass--through--able'' (dead end). A simulation trial always starts from a real--world situation in which the agent detects the existing obstacles, resulting in the initial sensory state $\vecs_{0}$. Afterwards, the simulation is performed by iteratively applying the IM and the visual and tactile FM. The IM suggest the next motor command, and the FMs predict the resulting new visual state $\widehat{\vecs}_{t+1}$ and the tactile state (see Fig.~\ref{fig_sim}). Afterwards the next iteration starts, this time based on the predicted visual state. In our experiments, the maximum number of iterations amounted to 60.

Whether an obstacle arrangement is ``pass--through--able'', is recognized based on the motor commands in the simulated sequence: If such a sequence is collision--free, contains only up to 20 turns, and the absolute difference between left and right turns does not exceed 10, an obstacle arrangement can be classified as corridor. Thus, the ``corridor criterion'' does not rely on sensory features, but on motor data gathered during the sensorimotor simulation. As soon as the corridor criterion is violated during a simulation trial, it is prematurely terminated.

So far, we have only considered a single simulation trial. In the following, we will introduce the additional term ``simulation run''. A simulation run consists of a number of simulation trials, all starting from the same real--world situation and initial sensory state $\vecs_{0}$. In our experiments, a simulation run consisted of up to 30 simulation trials. Since the output of the IM is deterministic in its basic version, it is necessary to inject some randomness into the simulation process to arrive at a different movement sequence in each trial (see Sects.~\ref{sect_invVar}/\ref{sect_invRestart}). If any of the trials within the run fulfills the corridor criterion, the obstacle arrangement is classified as corridor, otherwise as dead end. After a classification as corridor, the simulation run is prematurely terminated. By using several simulation trials instead of only one, the final classification is based on a more substantial amount of (internally generated) sensorimotor data.

\section{Experimental Results}\label{sect_exp}

\subsection{Task Conditions}

In the final experiments, the overall architecture was tested regarding its perceptual performance, i.e.~its ability to distinguish dead ends from corridors by internal simulation. In these experiments, we varied three aspects of the internal simulation process systematically: (1) the type of IM; (2) how to deal with oscillations (left turn followed by right turn followed by left turn$\dots$); (3) how to (re-)start a simulation trial.

\subsubsection{Variations of the Inverse Model}\label{sect_invVar}

We compared three different types of IMs: The IM as described in Sect.~\ref{sect_im} with deterministic output (called DET), a probabilistic interpretation of this IM (called PROB), and a random walk (called RANDOM).

In the probabilistic version, the outputs of the three regression modules are interpreted as  likelihoods that the corresponding motor commands are the optimum choice. First each value $\hat{q}(m_i,m_j)$ (with $m_1 = \mbox{forward}, m_2 = \mbox{left}, m_3 = \mbox{right}, i \neq j$) is clamped to the range $[0;1]$. Afterwards, intermediate (yet unnormalized) probabilities are computed:
\begin{eqnarray*}
\widetilde{p}(m_i) = \hat{q}(m_i,m_j) \cdot \hat{q}(m_i,m_k) \quad \mbox{with} \; i \neq j, i \neq k, j \neq k; i,j,k \in \{1,2,3\}
\end{eqnarray*}
Final probabilities are determined by normalizing the $\widetilde{p}(m_i)$ values:
\begin{eqnarray*}
p(m_i) = \frac{\widetilde{p}(m_i)}{\sum_{j=1}^{3}\widetilde{p}(m_j)}
\end{eqnarray*}
Based on the probabilities $p(m_i)$, the output of IM--PROB is randomly generated.

IM--RANDOM generates a pure random walk with the following basic probabilities: $p(\mbox{forward}) = 2/3, p(\mbox{left}) = 1/6, p(\mbox{right}) = 1/6$. In this way, the probabilities for the occurence of each motor command in the simulated sequences match the corridor criterion (a maximum number of 20 turns is allowed within a sequence of 60 steps). This is a necessary prerequisite for a fair comparison with the other types of IMs. The purpose of IM--RANDOM is to serve as baseline condition.

\subsubsection{Anti--Oscillation Mode}

Whenever an IM generates a turn which counteracts the turn directly before, this leads in case of IM--DET to a deadlock, and in case of IM--PROB and IM--RANDOM to frequent violations of the corridor criterion because the maximum number of allowed turns is more quickly exceeded. We tested four ways to counteract this problem (called anti--oscillation modes in the following). In the first mode (NONE), no countermeasures are taken. In the second mode (CONTINUE), we proceed as in the original study by \cite{moeller+08}: Whenever two subsequent turns cancel each other out, the second turn is changed to a turn into the same direction as the first one. In the third mode (FORWARD), both turns are deleted from the movement sequence (the second rotation undoes the first rotation; the predicted sensory state is kept consistent by executing the visual FM for the second turn). If a collision--free forward movement is now possible (tested with the FMs), this single forward movement is inserted into the movement sequence instead of the two deleted turns. Otherwise, the IM has to directly determine the next movement step. This leads in case of IM--DET to a deadlock; for this reason, a simulation trial is stopped after 300 invocations of the IM. The fourth mode (FORWARD--CONTINUE) works very similar to the third mode. After the deletion of both turns a forward movement is tested. If this is collision--free, it is inserted into the movement sequence, otherwise two equal turns in direction of the first deleted turn. This anti--oscillation mode basically combines the FORWARD and the CONTINUE strategy and avoids deadlocks.

\subsubsection{Restart Mode}\label{sect_invRestart}

Whenever a simulation trial is stopped because the movement sequence indicates a dead end or because the maximum iteration depth is reached, a new trial has to be started. In case of the restart mode "FULL", the whole movement sequence from the previous trial is discarded; the internal simulation starts again from scratch. This works well for IMs with a probabilistic component, while IM--DET produces always the same movement sequence. However, for completeness the combination of IM--DET with FULL was included in the experiments. The second restart mode is taken from the original study by \cite{moeller+08}. In this mode, each movement sequence is identical to the one from the previous trial up to a randomly selected step within the first two thirds of the sequence (therefore this mode is called PARTIAL). At that point, the new sequence is modified by performing $3$ subsequent rotations either to the left or to the right (random decision). For all subsequent steps, the internal simulation follows again the suggestions of the IM. By inserting random rotations, the PARTIAL restart mode injects randomness into the simulation process and is thus especially suited for IM--DET.

\subsection{Experimental Procedure}

Overall, they are 24 task conditions (3 types of IMs $\times$ 4 anti--oscillation modes $\times$ 2 restart modes). In each task condition, we carried out 600 simulation runs based on the robot's camera images of 60 different real--world obstacle arrangements; thus the image of every obstacle arrangement ist presented 10 times in each task condition as initial real--world state. The 60 obstacle arrangements are devided into 20 dead ends and 40 corridors. Each one consists of 10 obstacles which are distributed over an area of about 3 $\times$ 4 meters (see Figs.~\ref{fig_compare} and \ref{fig_invCompare} for some examples). The robot was placed close to the entrance of each dead end or corridor situation. We ensured that in every of these 60 scenarios all obstacles were detected in the image processing stage (however, specific properties like their width might have been misperceived, see Fig.~\ref{fig_deadendMiss}). Dead ends were arranged such that the maximum gap between obstacles amounted to 65 cm. In corridor scenarios, there were one (or rarely two) gaps with a minimum width of 85 cm. The maximum gap size was 110 cm.

\subsection{Results}

We report for each task condition the success rates (correctly identified corridors and dead ends in Tab.~\ref{tab_resErr}), the average number of simulation trials per run in case of successful corridor detection (Tab.~\ref{tab_resTrial}), and the overall number of FM invocations over all 600 simulation runs (Tab.~\ref{tab_resFwm}).

\subsubsection{Success Rates}

\begin{sloppypar}
The success rates in Tab.~\ref{tab_resErr} show that the cognitive architecture is well suited to reveal the affordances of obstacle arrangements --- at least in some task conditions. Dead ends are nearly always correctly classified, and corridor detection works well for IM--DET and IM--PROB if combined with the appropriate anti--oscillation mode (FORWARD/FORW.--CONTINUE) and the appropriate restart mode (PARTIAL for IM--DET, FULL for IM--PROB). In these task conditions, the success rates for corridor detection are about 90\%. The best result (91.3\%) is obtained from the combination IM--DET/FORW.--CONTINUE/PARTIAL. In comparison, the combination IM--DET/CONTINUE/PARTIAL which was applied by \cite{moeller+08} reaches only 65.8\%. This is surprising but illustrates the value of real--world experiments in complementing pure simulation studies.
Another unexpected result is the rather good performance of the random walk (IM--RANDOM): It reaches in combination with FORWARD/FULL a corridor success rate of 76.5\%. Furthermore, IM--RANDOM even finds passages through some dead end scenarios. A close inspection of these cases reveals that this is caused by misperceived obstacles. This misperception has already been discussed in Sect.~\ref{sect_improc_corr} and is illustrated in Fig.~\ref{fig_deadendMiss}. Even with the mentioned correction steps, the width and vertical offset of segments is sometimes wrongly estimated. In this case, artifical gaps may open in dead ends as shown in in Fig.~\ref{fig_deadendMiss} (see the caption of Fig.~\ref{fig_invCompare} on how the illustration from the bird's eye view was generated).
\end{sloppypar}

For probabilistic IMs, the FULL restart mode is generally better suited as the original PARTIAL strategy. Most likely, because by a full restart a bad movement sequence in the beginning is always completely discarded, and the occurence of such bad sequences might be more likely in case of probabilistic movement decisions. Comparing the anti--oscillation modes, the CONTINUE strategy performs nearly as bad as working without any strategy (NONE). The FORWARD and FORW.--CONTINUE strategies work both rather well; the FORWARD strategy is superior if combined with IM--RANDOM. This is most likely the case because it leaves more room for random decisions and the revision of previously executed bad decisions.

\subsubsection{Average Number of Trials}

The random walk was rather successful given the number of correctly classified scenarios which questions the necessity of sophisticated IMs. However, the average number of trials for successfully detected corridors in Tab.~\ref{tab_resTrial} shows clearly that this is a premature conclusion. In its most precise task condition FORWARD/FULL, IM--RANDOM needs on average 8.56 simulation trials, compared with approx.~4.5 required by IM--PROB and only about 2.5 required by IM--DET! This proves convincingly that sophisticated IMs in general and the deterministic variant in particular are far superior to a pure random walk.

\subsubsection{Average Number of FM Invocations}

In addition, we analyzed the average number of FM invocations over all 600 simulation runs (Tab.~\ref{tab_resFwm}). This measure also reflects how costly it is to generate movement sequences in case of dead ends. Again, the random walk is very expensive compared to the more sophisticated IMs (3698 invocations for IM--RANDOM/FORWARD/FULL in contrast to 929 invocations for IM--PROB/FORW.--CONTINUE/FULL and 686 invocations for IM--DET/FORW.--CONTINUE/PARTIAL). This further corroborates the impression that the success of random strategies depends on a large number of tested movement steps while sophisticated IMs are more efficient because their output is goal--directed.

\renewcommand{\mf}{\scriptsize}
\renewcommand{\arraystretch}{1.1}
\begin{table}
  \begin{tabularx}{\textwidth}{l|R|R|R|R|R|R}
    \hline
    \multicolumn{1}{c|}{} & \multicolumn{6}{c}{\mf \bf Inverse Model} \\
    \multicolumn{1}{c|}{} & \multicolumn{2}{c|}{\mf DET} & \multicolumn{2}{c|}{\mf PROB} & \multicolumn{2}{c}{\mf RANDOM} \\\cline{2-7}
    \multicolumn{1}{l|}{\mf \bf \rb{0.2cm}{Anti--Oscillation}} & \multicolumn{2}{c|}{\mf \bf Restart Mode} & \multicolumn{2}{c|}{\mf \bf Restart Mode} & \multicolumn{2}{c}{\mf \bf Restart Mode} \\
    \multicolumn{1}{l|}{\mf \bf Mode} & \mf FULL & \mf PARTIAL & \mf FULL & \mf PARTIAL & \mf FULL & \mf PARTIAL \\ \Hline
    \mf NONE & \mf 32.5 / 100.0 & \mf 65.5 / 100.0 & \mf 31.3 / 100.0 & \mf 12.8 / 100.0 & \mf 21.0 / 100.0 & \mf 5.3 / 100.0 \\ \hline
    \mf FORWARD & \mf 60.0 / 100.0 & \mf 90.0 / 100.0 & \mf 88.3 / 98.0 & \mf 76.5 / 100.0 & \mf 76.5 / 95.5 & \mf 49.3 / 100.0 \\ \hline
    \mf FORW.--CONTINUE & \mf 65.0 / 100.0 & \mf 91.3 / 100.0 & \mf 88.0 / 98.0 & \mf 71.3 / 100.0 & \mf 56.3 / 99.5 & \mf 32.0 / 100.0 \\ \hline
    \mf CONTINUE & \mf 45.0 / 100.0 & \mf 65.8 / 97.5 & \mf 27.0 / 100.0 & \mf 8.5 / 100.0 & \mf 27.5 / 100.0 & \mf 4.8 / 100.0 \\ \hline
  \end{tabularx}
  \caption{\label{tab_resErr} Success rates for detecting the right affordance in percent (values for corridors on the left in each cell, for dead ends on the right). }
\end{table}

\begin{table}
  \begin{tabularx}{\textwidth}{l|R|R|R|R|R|R}
    \hline
    \multicolumn{1}{c|}{} & \multicolumn{6}{c}{\mf \bf Inverse Model} \\
    \multicolumn{1}{c|}{} & \multicolumn{2}{c|}{\mf DET} & \multicolumn{2}{c|}{\mf PROB} & \multicolumn{2}{c}{\mf RANDOM} \\\cline{2-7}
    \multicolumn{1}{l|}{\mf \bf \rb{0.2cm}{Anti--Oscillation}} & \multicolumn{2}{c|}{\mf \bf Restart Mode} & \multicolumn{2}{c|}{\mf \bf Restart Mode} & \multicolumn{2}{c}{\mf \bf Restart Mode} \\
    \multicolumn{1}{l|}{\mf \bf Mode} & \mf FULL & \mf PARTIAL & \mf FULL & \mf PARTIAL & \mf FULL & \mf PARTIAL \\ \Hline
    \mf NONE & \mf 1.00 (0.00) & \mf 3.82 (4.79) & \mf 8.14 (7.87) & \mf 6.37 (8.14) & \mf 13.73 (8.55) & \mf 10.95 (8.20) \\ \hline
    \mf FORWARD & \mf 1.00 (0.00) & \mf 2.36 (3.28) & \mf 4.59 (5.75) & \mf 5.03 (6.99) & \mf 8.56 (7.29) & \mf 8.27 (8.18) \\ \hline
    \mf FORW.--CONTINUE & \mf 1.00 (0.00) & \mf 2.45 (3.85) & \mf 4.26 (5.08) & \mf 5.08 (7.02) & \mf 11.28 (7.85) & \mf 11.08 (8.65) \\ \hline
    \mf CONTINUE & \mf 1.00 (0.00) & \mf 2.89 (4.21) & \mf 9.76 (8.43) & \mf 8.00 (7.62) & \mf 13.99 (8.99) & \mf 13.68 (9.25) \\ \hline
  \end{tabularx}
  \caption{\label{tab_resTrial} Average number of trials over all successful simulation runs on corridors (standard deviation given in brackets). The number of successful runs depends on the task condition.}
\end{table}

\begin{table}
  \begin{tabularx}{\textwidth}{l|R|R|R|R|R|R}
    \hline
    \multicolumn{1}{c|}{} & \multicolumn{6}{c}{\mf \bf Inverse Model} \\
    \multicolumn{1}{c|}{} & \multicolumn{2}{c|}{\mf DET} & \multicolumn{2}{c|}{\mf PROB} & \multicolumn{2}{c}{\mf RANDOM} \\\cline{2-7}
    \multicolumn{1}{l|}{\mf \bf \rb{0.2cm}{Anti--Oscillation}} & \multicolumn{2}{c|}{\mf \bf Restart Mode} & \multicolumn{2}{c|}{\mf \bf Restart Mode} & \multicolumn{2}{c}{\mf \bf Restart Mode} \\
    \multicolumn{1}{l|}{\mf \bf Mode} & \mf FULL & \mf PARTIAL & \mf FULL & \mf PARTIAL & \mf FULL & \mf PARTIAL \\ \Hline
    \mf NONE & \mf 1144 (612) & \mf 777 (524) & \mf 1359 (539) & \mf 1236 (356) & \mf 1587 (375) & \mf 1304 (185) \\ \hline
    \mf FORWARD & \mf 2936 (5217) & \mf 1406 (2378) & \mf 1338 (1550) & \mf 1256 (1130) & \mf 3698 (2436) & \mf 2864 (1433) \\ \hline
    \mf FORW.--CONTINUE & \mf 967 (870) & \mf 686 (663) & \mf 929 (774) & \mf 1047 (731) & \mf 1795 (785) & \mf 1525 (490) \\ \hline
    \mf CONTINUE & \mf 1011 (690) & \mf 641 (457) & \mf 1225 (436) & \mf 1037 (248) & \mf 1523 (408) & \mf 1293 (171) \\ \hline
  \end{tabularx}
  \caption{\label{tab_resFwm} Average number of FM invocations over all 600 simulation runs (400 corridors, 200 dead ends) (standard deviation given in brackets).}
\end{table}

Generally, the anti--oscillation modes CONTINUE and FORW.--CONTINUE are the most economic ones while FORWARD leads in contrast to the largest number of FM invocations. For this reason, the combination IM--PROB/FORW.--CONTINUE/FULL is more appealing than IM--PROB/FORWARD/FULL although the latter has a slightly better success rate. However, in the end the deterministic version of the IM is the most preferable one. Combined with FORW.--CONTINUE/PARTIAL it shows consistently very good results in all considered measures. The striking difference to the previous pure simulation study \cite[]{moeller+08} is that the pure CONTINUE mode does not result in good success rates for corridor detection.

\begin{figure}[p]
\begin{center}
\begin{tabular}{c|c|c}
\large IM--DET & \large IM--PROB & \large IM--RANDOM \\[0.2cm] \hline
% p3/deadend01
\includegraphics[width=0.3\textwidth]{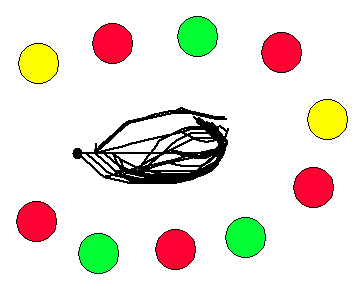} &
\includegraphics[width=0.3\textwidth]{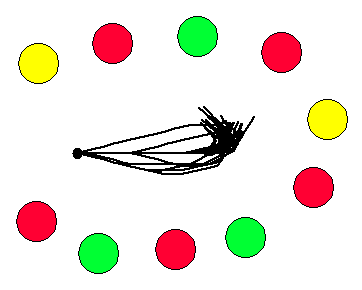} &
\includegraphics[width=0.3\textwidth]{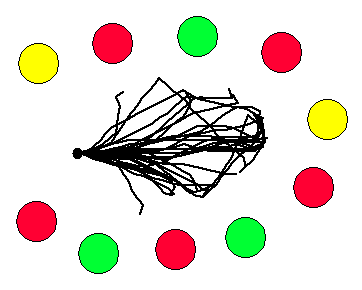} \\[0.1cm]\hline
% p3/passage02a
\includegraphics[width=0.3\textwidth]{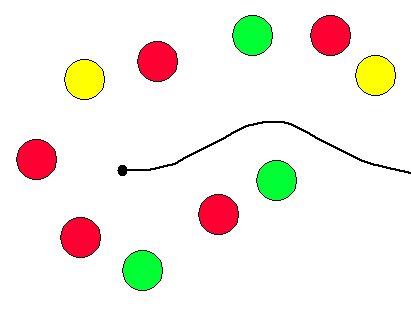} &
\includegraphics[width=0.3\textwidth]{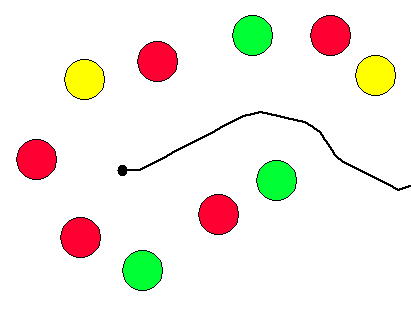} &
\includegraphics[width=0.3\textwidth]{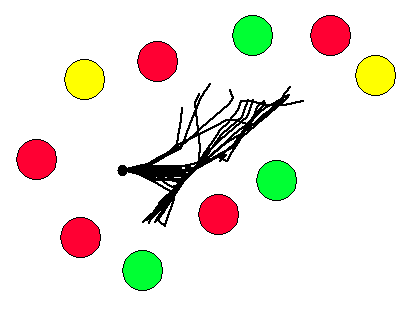} \\[0.1cm]\hline
% p3/passage06b
\includegraphics[width=0.3\textwidth]{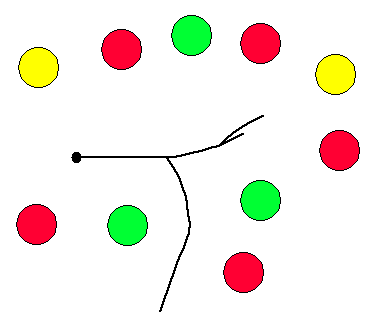} &
\includegraphics[width=0.3\textwidth]{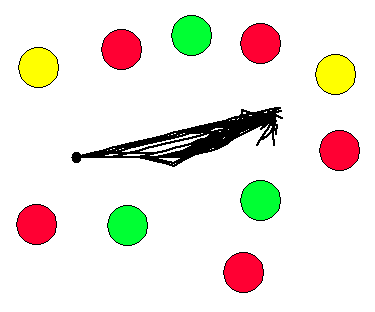} &
\includegraphics[width=0.3\textwidth]{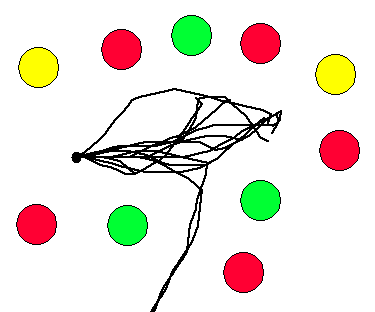} \\[0.1cm]\hline
% p3/passage13b
\includegraphics[width=0.3\textwidth]{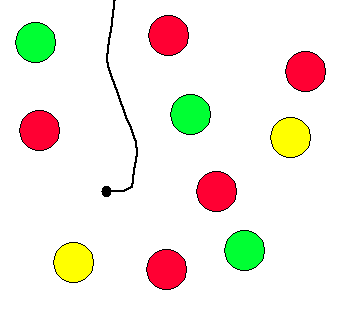} &
\includegraphics[width=0.3\textwidth]{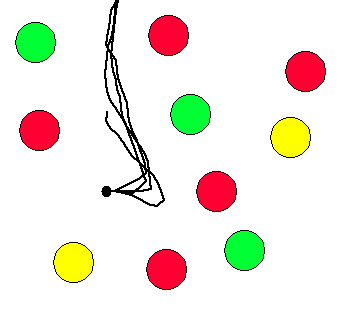} &
\includegraphics[width=0.3\textwidth]{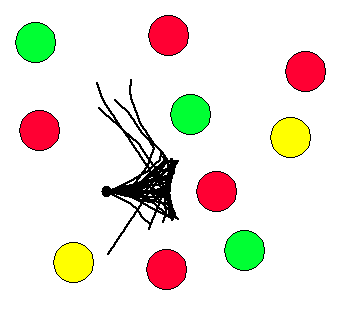} \\[0.1cm]\hline
\end{tabular}
\caption{\label{fig_invCompare}\small Illustration of several obstacle arrangements and the movement sequences generated by the different types of inverse models (IM--DET combined with FORW.--CONTINUE/PARTIAL, IM--PROB with FORW.--CONTINUE/FULL, and IM--RANDOM with FORWARD/FULL) in single simulation runs. The top row shows a dead end situation, the rows below corridors. IM--PROB fails in the third row, IM--RANDOM in the second and fourth row. Note that these illustrations from a bird's eye perspective are not part of the cognitive architecture; they were only generated for illustration purposes. The location of the obstacles is determined from the estimated horizontal offset $x$ and width $w$ of the corresponding segment (after the correction steps in Sect.~\ref{sect_improc_corr}). Thus, the depicted obstacle locations reflect the ``perception'' of the robot and not their real locations. To compute the distance on the ground between robot and obstacle from the width $w$ a series of additional calibration images was recorded which were not used otherwise in any part of the cognitive architecture.}
\end{center}
\end{figure}

\subsubsection{Visualization}

Fig.~\ref{fig_invCompare} illustrates the output of the different types of IMs from a bird's eye perspective for one dead end and three corridors. Admittedly, these examples were purposefully picked by the authors, but they show nevertheless the different characteristics of the generated movement sequences.

IM--RANDOM produces many different sequences within a simulation run which cover a wide area. Nevertheless, rather obvious exits as in the second row of images are missed. In contrast, IM--DET produces gently curved movement sequences, and sometimes even a single one suffices to find the passage through a corridor. The characteristics of IM--PROB are in between. In the third row it does not identify the passage despite many attempts, most likely because the probability of a right turn is far too low at the crucial position within the arrangement. Here, the random injection of three subsequent right rotations by the PARTIAL strategy helps IM--DET to find the exit very quickly (of course, this is a lucky incident in this case).

\section{Discussion}\label{sect_concl}

Within its practical and technical limits, the proposed architecture enables the robotic agent to successfully recognize the behavioral meaning of obstacle arrangements as either dead ends or corridors. This is a non--trivial task, as Fig.~\ref{fig_compare} illustrates: For an outside observer, the difference between dead end and corridor is obvious (images in first row); however, from the robot's perspective (images in second and third row) this distinction is not obvious at all. The crucial step for the solution of this recognition problem is the transformation of the \emph{sensory} task into a \emph{sensorimotor} task by internal simulation based on mental imagery.
It is remarkable that during this transition a proximal sense (touch) starts to serve as basis for the interpretation of the data from a distal sense (vision). The internal simulation is a ``projection'' from the distal visual sense onto the proximal tactile sense and onto motor commands and from there onto the assessment signals (``collisions are bad''), and this projection is what ultimately provides behavioral meaning to the visual image.

The hierarchical approach --- learning prediction by the forward models first, learning motor control by the inverse model afterwards --- is a very important aspect of the architecture. The way in which the inverse model is acquired can partly be characterized as mental training in which the mental images are produced by the already trained forward model.
The general principle ``prediction before control'' is consistent with results from psychological experiments \cite[e.g.,][]{flanagan+03}. In the final internal simulation for affordance detection, the inverse model plays an important role to guide the generation of movement sequences. Thus, the motor competence of the agent is an important prerequisite for its perceptual competence \cite[see][for corresponding results from psychology]{witt+08}.

The study in this paper builds upon our previous work \cite[]{moeller+08} in which we developed the cognitive architecture and tested it in a pure simulation study. In \cite{moeller+08}, we already discussed the basic design decisions and insights for cognitive science in--depth. We won't repeat this here with the same comprehensiveness, instead we will focus on the differences and on the insights from the new experiments. First of all, the transfer to a real--world setup is a considerable step forward on our way from a ``computerized thought experiment'' to real applications of the ``perception through anticipation'' (PtA) approach. Working with real--world data poses some additional challenges, in our case especially noise and uncertainties in the feature detection process. As stated earlier, each single obstacle is interpreted as a feature of the whole obstacle arrangement. In the sensory domain, these features appear as segments with specific properties like width and position. However, because of noisy image data and partial occlusions these parameters are sometimes wrongly registered. Errors like this can even change the behavioral meaning of the whole obstacle arrangement (i.e., a dead end becomes a corridor and vice versa). For this reason, we had to introduce an additional correction step in the feature acquisition process which is based on the overall distribution statistics of the parameters. This correction is not perfect but yields much better results in affordance detection than without (result from pre--tests). The general lesson from this is that the quality of the internal simulations depends on the precise registration of the initial real--world state, and that this precision can be enhanced by exploiting redundancy and by reconstructing incomplete features through implicit knowledge about typical feature characteristics --- without abandoning the basic principle of the PtA approach that pure sensory processing has to be kept at a rather simple level.

Furthermore, we extended our previous work \cite[]{moeller+08} by comparing different types of inverse models and other aspects of the simulation process in a systematic experimental study. The most important results can be summarized as follows: Although random walks can be rather successful in generating useful movement sequences, this comes at the price of a considerably larger simulation effort (required number of simulation trials, invocations of the forward model) compared to purposefully trained inverse models. This shows that the inverse model is not a dispensable part of the cognitive architecture, but instead a very important one (even more so if one considers the possibility of a hierarchy over several forward and inverse models at more and more abstract processing levels). For the injection of randomness into the simulation process, we compared two different methods: (1) Using a deterministic inverse model whose output is from time to time complemented by some random motor commands, or (2) by interpreting the output of the inverse model in a probabilistic way. In our view, the latter method is more elegant, however, the results show that the first method is superior (performance--wise only slightly, but regarding simulation effort rather significantly). Although this finding is most likely task--dependent, the general lesson from this is that the question on how to create some random variety in the simulation process deserves high attention because it can have a strong impact on the efficiency.

A third difference concerns the implementation of the forward model. In our previous work \cite[]{moeller+08}, we used a set of multi--layer perceptrons to approximate the training data. In the real--world setting with more noisy training data, an approach with a very strong regularization was more successful: fitting analytical functions with only a very small number of free parameters to the data. On the downside, this caused undesired extrapolation behavior which was counteracted by projecting the predicted parameters of the obstacle segments back onto the distribution of these parameters in the training data. This is conceptually similar to the approach by \cite{hoffmann07} who projected patches of predicted images back onto the distribution of these patches by a Gaussian mixture model (also for the purpose of avoiding ``catastrophic extrapolation'' in the internal multi--step simulation). We assume that such a predictor--corrector scheme is generally very useful to stabilize long--term simulations because it can partly compensate inprecise predictions.

In this way the present study identifies several important principles which can be used to successfully create robot implementations of computational models which are based on the concept of internal simulation \cite[e.g., simulation theories of perception and cognition; ][]{hesslow02,cruse03b,grush04}. With regard to its conceptual underpinnings and basic methodology, our work belongs to the fields of ``embodied cognition'' and ``developmental robotics'' \cite[]{lungarella+03,pfeifer+04} in that it aims on a sensorimotor foundation of cognition and puts a strong emphasis on learning. The final goal of the PtA approach is to explain the direct perception of more complex affordances in real--world settings (e.g., ``sit--on--able'' for a humanoid robot, or ``bimanually--graspable'' for a setup with two robot arms). One day affordance detection could complement or maybe even replace classical object recognition algorithms which work solely on sensory data. The presented research is a further advancement in this direction.

%---------------------------------------------------------------------------
% REFERENCES
%---------------------------------------------------------------------------

\bibliographystyle{apalike}
%\bibliography{/homes/wschenck/myDocuments/litFiles/CRlitlist,/homes/wschenck/myDocuments/dissertation/dissMain/dissMain,/homes/wschenck/myDocuments/litFiles/WSlitlist}

\end{document}